\documentclass[10pt,logo,copyright]{nvidiatechreport}
\usepackage[authoryear,round]{natbib}

\usepackage[utf8]{inputenc} 
\usepackage[T1]{fontenc}    

\usepackage{parskip}        
\usepackage{url}            
\usepackage{booktabs}       
\usepackage{amsfonts}       
\usepackage{nicefrac}       
\usepackage{microtype}      
\usepackage{xcolor}         
\usepackage[dvipsnames]{xcolor} 
\usepackage{graphicx}
\usepackage{animate}        
\usepackage{subcaption}
\usepackage{tabularx}
\usepackage{makecell}
\usepackage{adjustbox}
\usepackage{setspace}
\newcolumntype{M}[1]{>{\centering\arraybackslash}m{#1}}
\usepackage{float}
\usepackage{tikz}
\usetikzlibrary{positioning,shapes,arrows, decorations.pathreplacing, backgrounds}
\usepackage{amsmath,amsfonts,bm, bbm,leftindex}
\usepackage{multirow}
\usepackage{comment}
\usepackage{gensymb}
\usepackage{lipsum}
\usetikzlibrary{arrows.meta, positioning, fit}
\usepackage[para]{threeparttable}
\usetikzlibrary{tikzmark}

\usepackage[most]{tcolorbox}
\usepackage{fancyvrb}
\usepackage{fvextra}
\usepackage{dashrule}
\usepackage{nicematrix}

\usepackage{multicol}

\definecolor{joncolor}{RGB}{0,82,156} 

\definecolor{seungcolor}{RGB}{156,82,100}

\def\eqref#1{equation~\ref{#1}}

\def\1{\bm{1}}

\DeclareMathAlphabet{\mathsfit}{\encodingdefault}{\sfdefault}{m}{sl}
\SetMathAlphabet{\mathsfit}{bold}{\encodingdefault}{\sfdefault}{bx}{n}

\newcommand{\E}{\mathbb{E}}

\makeatletter
\let\save@mathaccent\mathaccent
\newcommand*\if@single[3]{%
  \setbox0\hbox{${\mathaccent"0362{#1}}^H$}%
  \setbox2\hbox{${\mathaccent"0362{\kern0pt#1}}^H$}%
  \ifdim\ht0=\ht2 #3\else #2\fi
  }
\newcommand*\rel@kern[1]{\kern#1\dimexpr\macc@kerna}
\newcommand*\widebar[1]{\@ifnextchar^{{\wide@bar{#1}{0}}}{\wide@bar{#1}{1}}}
\newcommand*\wide@bar[2]{\if@single{#1}{\wide@bar@{#1}{#2}{1}}{\wide@bar@{#1}{#2}{2}}}
\newcommand*\wide@bar@[3]{%
  \begingroup
  \def\mathaccent##1##2{%
    \let\mathaccent\save@mathaccent
    \if#32 \let\macc@nucleus\first@char \fi
    \setbox\z@\hbox{$\macc@style{\macc@nucleus}_{}$}%
    \setbox\tw@\hbox{$\macc@style{\macc@nucleus}{}_{}$}%
    \dimen@\wd\tw@
    \advance\dimen@-\wd\z@
    \divide\dimen@ 3
    \@tempdima\wd\tw@
    \advance\@tempdima-\scriptspace
    \divide\@tempdima 10
    \advance\dimen@-\@tempdima
    \ifdim\dimen@>\z@ \dimen@0pt\fi
    \rel@kern{0.6}\kern-\dimen@
    \if#31
      \overline{\rel@kern{-0.6}\kern\dimen@\macc@nucleus\rel@kern{0.4}\kern\dimen@}%
      \advance\dimen@0.4\dimexpr\macc@kerna
      \let\final@kern#2%
      \ifdim\dimen@<\z@ \let\final@kern1\fi
      \if\final@kern1 \kern-\dimen@\fi
    \else
      \overline{\rel@kern{-0.6}\kern\dimen@#1}%
    \fi
  }%
  \macc@depth\@ne
  \let\math@bgroup\@empty \let\math@egroup\macc@set@skewchar
  \mathsurround\z@ \frozen@everymath{\mathgroup\macc@group\relax}%
  \macc@set@skewchar\relax
  \let\mathaccentV\macc@nested@a
  \if#31
    \macc@nested@a\relax111{#1}%
  \else
    \def\gobble@till@marker##1\endmarker{}%
    \futurelet\first@char\gobble@till@marker#1\endmarker
    \ifcat\noexpand\first@char A\else
      \def\first@char{}%
    \fi
    \macc@nested@a\relax111{\first@char}%
  \fi
  \endgroup
}
\makeatother

\usepackage{pifont}
\usepackage{diagbox}
\usepackage{graphicx}

\usepackage[nameinlink]{cleveref}
\crefname{equation}{Eq.}{Eqs.}
\crefname{figure}{Fig.}{Figs.}
\crefname{section}{Sec.}{Sec.}
\crefname{appendix}{App.}{App.}
\crefname{table}{Tab.}{Tabs.}
\crefname{algorithm}{Algo}{Algo}
\crefname{thm}{Thm}{Thm}
\Crefname{thm}{Thm}{Thm}
\crefname{prop}{Prop}{Prop}

\definecolor{darkred}{rgb}{0.7, 0.0, 0.0}

\newcommand{\crefnames}[3]{%
  \@for\next:=#1\do{%
    \expandafter\crefname\expandafter{\next}{#2}{#3}%
  }%
}
\newcommand{\modelname}{\emph{OmniDreams}\xspace}
\newcommand{\modelnamesv}{\emph{OmniDreams-SV}\xspace}
\newcommand{\modelnamemv}{\emph{OmniDreams-MV}\xspace}

\title{NVIDIA OmniDreams: Real-Time Generative World Model for Closed-Loop Autonomous Vehicle Simulation}

\author{NVIDIA\footnote{A detailed list of contributors and acknowledgments can be found in~\cref{sec::contributors} of this paper.}}

\hypersetup{
  pdftitle    = {NVIDIA OmniDreams: Real-Time Generative World Model for Closed-Loop Autonomous Vehicle Simulation},
  pdfauthor   = {NVIDIA Spatial Intelligence Lab},
  pdfsubject  = {Real-time generative closed-loop autonomous vehicle simulation},
  pdfkeywords = {NVIDIA OmniDreams, autonomous vehicles, world models, closed-loop simulation, video diffusion, NVIDIA, Cosmos, Alpamayo, AlpaSim},
  pdfcreator  = {LaTeX with hyperref},
}

\begin{abstract}

As autonomous vehicle (AV) capabilities advance, the safe evaluation of driving policies in long-tail scenarios remains a critical bottleneck. In closed-loop simulation, the driving policy model actively interacts with the environment, where its actions dynamically update the simulator state and directly influence the next set of generated sensor observations.
While recent reconstruction-based neural simulators offer photorealism, they are fundamentally constrained by their initial captured data and struggle to generalize to highly dynamic or novel scenes. 
To overcome these limitations, we introduce \modelname, a foundation generative world model mid- and post-trained from the Cosmos diffusion model to autoregressively generate action-conditioned videos in real time. 
By leveraging the rich visual priors of Cosmos and mid- and post-training on 21k hours of driving scenarios, \modelname synthesizes complex, unobserved phenomena that are hard for traditional simulators to capture, such as extreme weather and unpredictable dynamic agent behaviors. 
Crucially, it autoregressively conditions its photorealistic sensor generation on past frames, the current simulator state, and immediate driving actions. 
Deployed in a closed-loop system with the Alpamayo 1 policy model and AlpaSim orchestrator, \modelname acts as a highly responsive, reactive environment, providing a scalable and comprehensive solution for training and evaluating next-generation autonomous driving policies. 
We additionally show preliminary results indicating that a world-action model (WAM) post-trained from OmniDreams achieves strong performance on the Physical AI Autonomous Vehicles NuRec dataset, surpassing the VLA-based Alpamayo 1.5 research policy model while using only 1/5 the total parameters. These results highlight the potential for a real-time world model like OmniDreams to also serve as a backbone for policy architectures. The code is available at \href{https://github.com/nv-tlabs/omni-dreams}{OmniDreams Github page}, and model weight at \href{https://huggingface.co/nvidia/omni-dreams-models}{Hugging face}. 

\end{abstract}

\begin{document}
\maketitle
\abscontent

\newpage
\tableofcontents
\newpage

\section{Introduction}
\label{sec::intro}

\begin{figure}[t]
    \centering
    \includegraphics[width=0.95\linewidth]{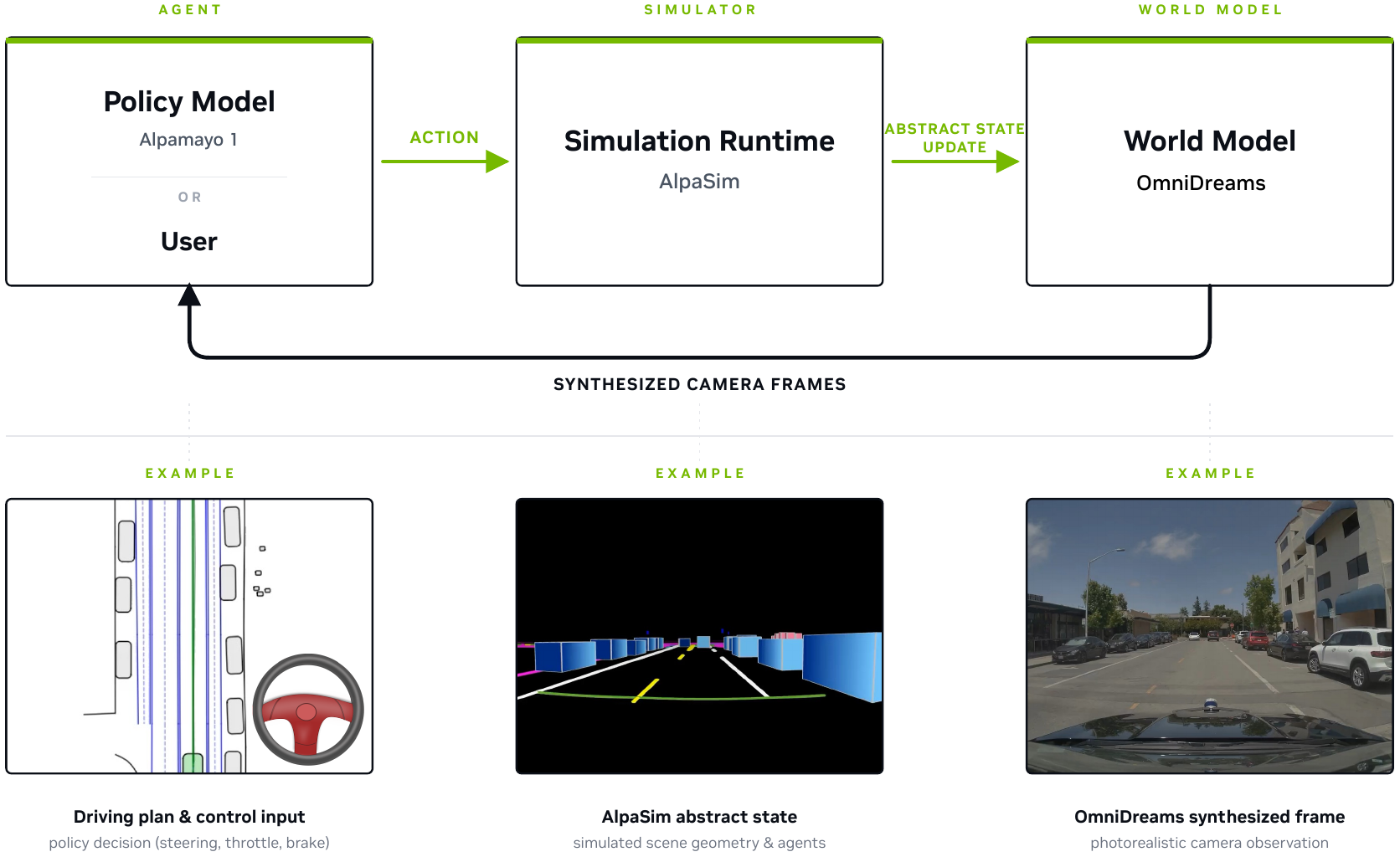}
    \caption{Closed-loop simulation workflow. A policy model (here, Alpamayo 1~\citep{nvidia2026alpamayo}) or user sends an action to the AlpaSim simulation runtime. AlpaSim updates the simulation state and forwards the context to \modelname, which synthesizes the next camera frames and returns them to the policy, completing the loop.}
    \label{fig:teaser}
\end{figure}

Autonomous vehicle (AV) research is entering a new phase. Reasoning-capable vision–language–action (VLA) models can ``think through'' complex situations, producing interpretable reasoning traces alongside planned trajectories~\citep{nvidia2026alpamayo}. Yet the core challenge remains: how to test these policies in safety-critical, long-tail scenarios before real-world deployment.
Meeting this challenge requires a simulator that is both interactive and scalable. In closed-loop evaluation, the policy actively drives within the simulator: it produces an action, the environment updates accordingly, and the policy receives the next sensor observations rendered from the updated state, shown in \cref{fig:teaser}. This is essential because the policy actions influence how a scenario unfolds over time.

Recent progress in reconstruction-based neural simulation has significantly advanced AV development~\citep{nvidia2024nurec}. By reconstructing real-world scenes from captured data, these systems enable photorealistic what-if testing within environments grounded in reality. As a result, developers can modify agent behaviors or trajectories and evaluate how policies respond under controlled variations of real scenarios.
However, reconstruction-based workflows remain fundamentally anchored to the data that was originally observed. While they support many what-if scenarios within a reconstructed scene, they struggle to scale beyond the captured corridor, introduce new scene content, or generate consistent photorealistic observations under substantially new conditions.

In contrast, generative world models trained on massive amounts of visual data learn rich priors about how the world behaves and evolves visually. These priors allow such models to synthesize highly challenging phenomena, such as highly dynamic weather conditions (e.g., rain, storms, snow, and wind) or unusual, deformable objects (e.g., a mattress on top of a car)~\citep{ren2025cosmosdrivedreams}. They can also generate complex agent behaviors, including pedestrian behavior and articulated motion that rapidly change in response to the vehicle’s proximity or stopping behavior.

This technical report introduces NVIDIA's latest research on closed-loop simulation via a generative world foundation model for autonomous driving. At the core of this research is a new foundation model, \modelname, an action-conditioned generative world model that is mid- and post-trained on Cosmos~\citep{nvidia2025worldsimulationvideofoundation}. \modelname generates photorealistic sensor observations while remaining interactive and responsive to the driving policy.

\modelname includes the following key design concepts behind the workflow:
\begin{itemize}
\item operating as an autoregressive diffusion video generation-based simulator, which simulates in a closed-loop with the open-sourced Alpamayo 1~\citep{nvidia2026alpamayo} as the policy model and NVIDIA AlpaSim~\citep{nvlabs2025alpasim} as the simulation orchestrator 
\item the KV cache stored from past generations is also attended during new frame generation, allowing \modelname to achieve long rollout consistency
\item conditioning sensor generation on simulator state and driving actions
\item real-time interactive rendering. The single-camera version of the 2B \modelname can render 68\,FPS videos at 720p resolution ($704 \times 1280$) with one GB300 GPU, whereas the 4-camera version of the 2B \modelname can render up to 105\,FPS at 720p resolution with 16 GB300 GPUs
\end{itemize}

Furthermore, we demonstrate that the internal representations learned by \modelname are inherently valuable for downstream driving tasks. Specifically, we show some preliminary results that a policy model post-trained from \modelname improves performance on the PAI autonomous driving dataset, reducing \texttt{collision} from $6.9\%$ to $4.2\%$,
\texttt{collision\_front} from $1.0\% \to 0.9\%$, \texttt{collision\_lateral} from $0.6\% \to 0.4\%$ and \texttt{collision\_rear} from $5.3\% \to 3.0\%$. 
Remarkably, it achieves this despite requiring roughly five times fewer parameters ($\sim$2\,B vs.\ $\sim$10\,B) than its VLA counterpart, Alpamayo 1.5~\citep{nvidia2026alpamayo}. To the best of our knowledge, this is one of the first works to show that World-Action Models (WAMs) can offer a superior solution to VLAs in autonomous driving, echoing recent findings in robotics~\citep{ye2026world}.

\section{Data}\label{sec::data}
Training a closed-loop generative world model places specific demands on the data pipeline: the model must learn to synthesize photorealistic, temporally consistent sensor observations conditioned on structured scene representations, ego trajectories, and textual environment descriptions. This section describes the data sources, conditioning signal extraction, curation process, and final dataset composition used to mid- and post-train \modelname from the Cosmos-Predict 2.5 foundation model~\citep{nvidia2025worldsimulationvideofoundation}.

\subsection{Data Sources}\label{sec::data_sources}
We use two AV datasets for training, Real Driving Scene (RDS) dataset used in~\citep{agarwal2025cosmos} but containing seven camera views, and RDS-HQ-1M, a newly curated and expanded version of RDS-HQ in~\citep{ren2025cosmosdrivedreams} with $1.14M$ clips. 
Both datasets are sourced from real-world driving logs under diverse conditions in 15 countries in Europe, Asia, and the US. They consist of 1080p video clips recorded with 7 synchronized cameras (front-wide, front-telescope, front-left, front-right, rear-left, rear-right, rear-tele) at 30 frames per second. RDS contains 3M 20s clips, selected to balance several driving scenarios, e.g., rural vs. city environments, daytime/nighttime scenes, and slow/fast driving. We use this dataset for mid-training, and it is also included in the Cosmos-Predict 2.5-AV pre-training set. 
RDS-HQ-1M contains 1.14M clips, a mix of 10s/20s scenes, with a focus on quality and highly accurate world-scenario annotations (see below). We use this dataset for finetuning and post-training.

\subsection{Conditioning Signal Extraction}\label{sec::conditioning}
As described in \cref{sec::model}, \modelname conditions generation on three input modalities: an abstract state of world-scenario map comprising rendered lane lines and bounding boxes which encodes the future ego trajectory, a text prompt describing environmental conditions, and a memory cache of recent visual history. Preparing paired training data, therefore, requires extracting these structured representations from raw driving logs.

\begin{figure*}[t]
    \centering
    \begin{subfigure}[t]{0.24\textwidth}
        \centering
        \includegraphics[width=\linewidth]{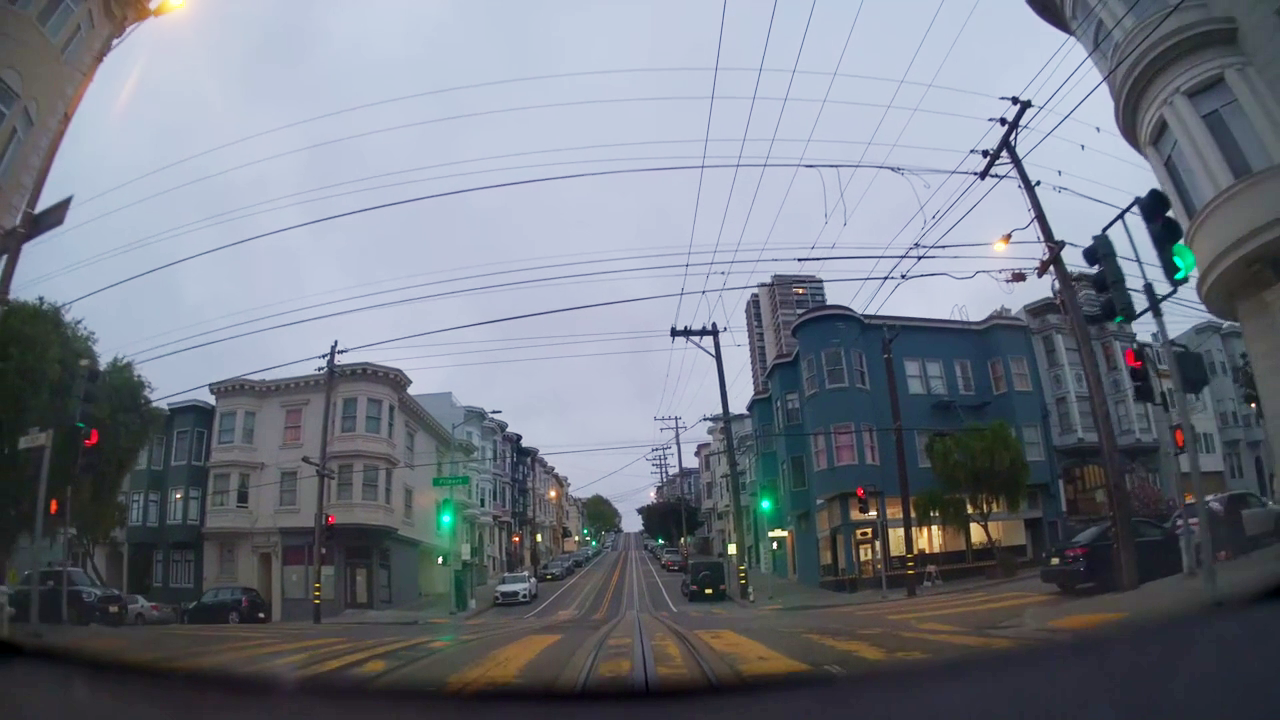}
        \caption{\centering{Front-wide camera}}
    \end{subfigure}
    \hfill
    \begin{subfigure}[t]{0.24\textwidth}
        \centering
        \includegraphics[width=\linewidth]{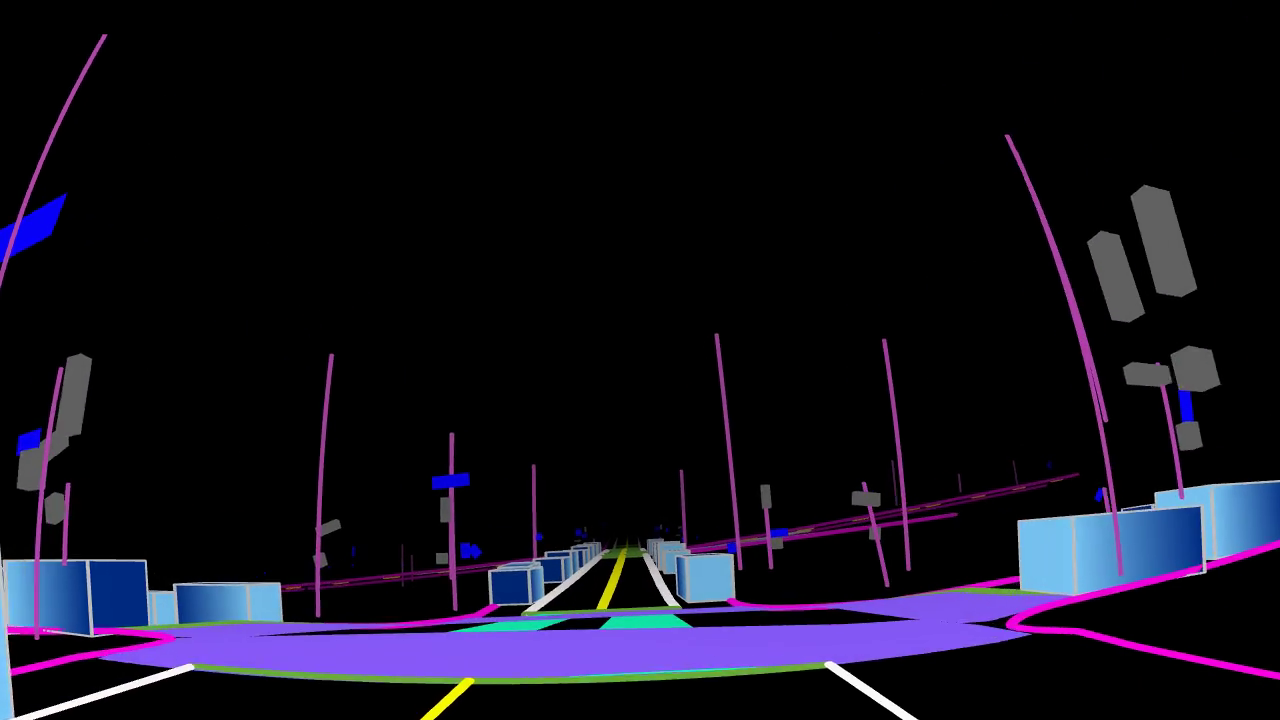}
        \caption{\centering{Front-wide world scenario (abstract state)}}
    \end{subfigure}
    \hfill
    \begin{subfigure}[t]{0.24\textwidth}
        \centering
        \includegraphics[width=\linewidth]{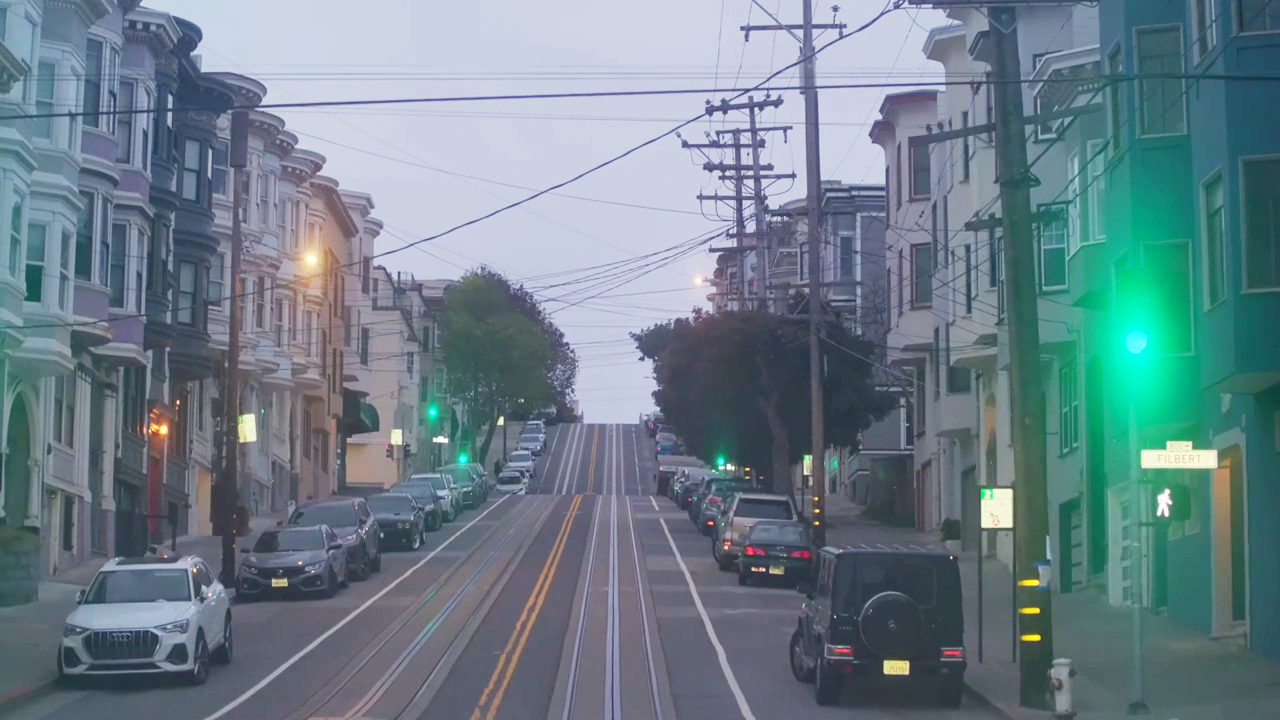}
        \caption{\centering{Rear-tele camera}}
    \end{subfigure}
    \hfill
    \begin{subfigure}[t]{0.24\textwidth}
        \centering
        \includegraphics[width=\linewidth]{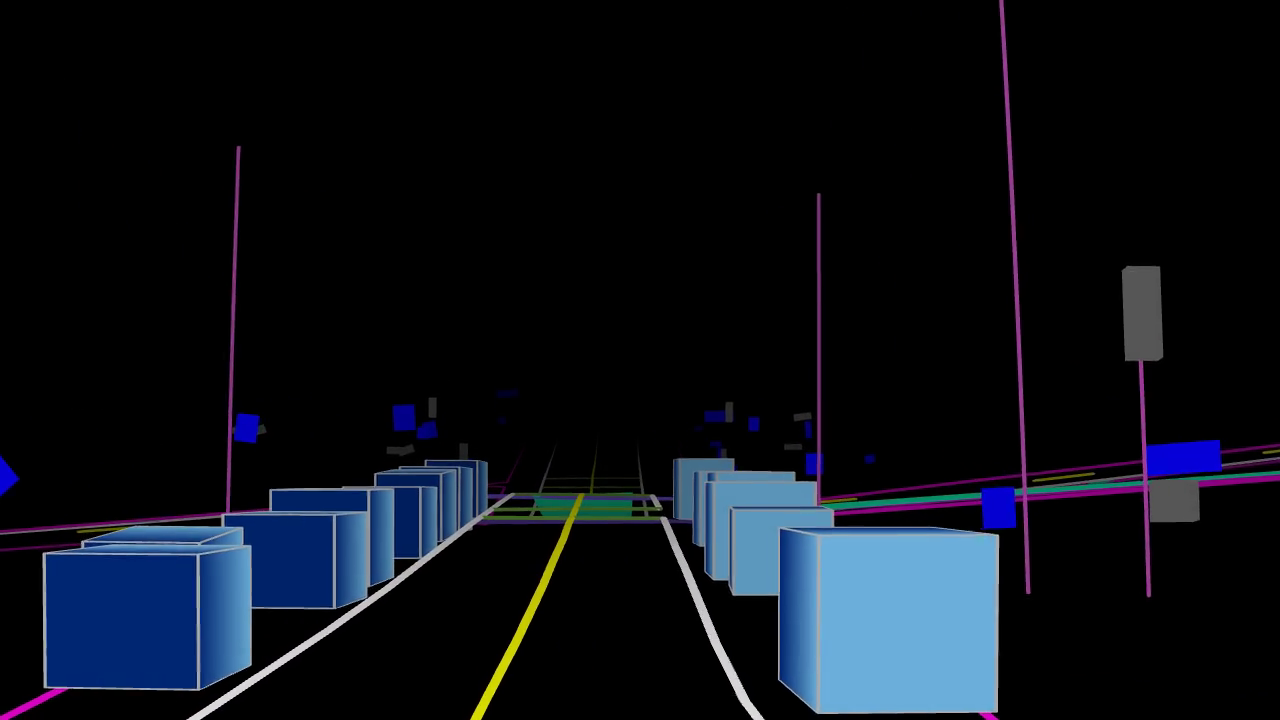}
        \caption{\centering{Rear-tele world scenario (abstract state)}}
    \end{subfigure}
    \vspace{-2mm}
    \caption{Example for world-scenario map rendering: lane lines (yellow), crosswalks (purple), cars (blue cuboids), and signs/signals (gray).}
    \label{fig:hdmap_example}
\end{figure*}

\paragraph{World-scenario map.}
The world-scenario map (abstract state) encodes both the high-definition map of the static scene and the dynamic actors that populate it. It also encodes the actions from the policy model or the user driver (in our real training data, actions come from test drivers).
High-definition (HD) maps provide detailed and precise road annotations, including lane lines, road boundaries, stop lines, poles, crosswalks, road markings, traffic lights, and traffic signs. All map labels are obtained from a pre-constructed city-level map and strictly matched to individual driving routes. We further perform 3D bounding-box detection and tracking of dynamic objects in each video clip, yielding 3D annotations for vehicles, pedestrians, and other vulnerable road users. 3D detection and tracking is run at 10\,FPS and interpolated to 30\,FPS to match the video frame rate. 

Then we follow the world-scenario map rendering scheme introduced in~\citep{nvidia2025worldsimulationvideofoundation}. For each camera view, we render these elements as pixel-aligned lines, surfaces, and 2D-projected cuboids using camera intrinsics and extrinsics. The cuboids for road actors are shaded to distinguish heading and provide additional depth cues. An example world-scenario rendering is shown in \cref{fig:hdmap_example}.

\paragraph{Text prompts.}
Text prompts enable control over environmental appearance such as weather, lighting, and time of day without modifying the underlying scene geometry. To generate text captions for training, we use a VLM (Qwen2.5-VL-7B~\citep{bai2023qwen}) to describe the video clips in our training sets. The VLM is tasked to describe the overall scenery, weather, driving behavior, and traffic. We perform captioning on 10s-long temporal windows and independently caption all 7 cameras. For each window, we generate three captions of varying lengths: short ($\approx$40 words), medium ($\approx$80 words), and long ($\approx$200 words). These captions are sampled with probabilities 0.1/0.2/0.7 during model training to make the model robust to prompt length.

As an example, this is one medium-length caption: \emph{``The video depicts a nighttime drive along a well-lit urban road. The ego vehicle moves steadily forward, passing under bright streetlights that illuminate the surroundings. The road features clear white lane markings, including directional arrows indicating turns. Traffic signs, including a yellow circular sign and a blue rectangular sign, are visible at intervals, providing guidance for drivers. The environment includes patches of greenery, bushes, and trees lining the roadside, adding depth to the scene. A billboard advertisement is also seen on the right side of the road. The road appears dry and in good condition.''}

\subsection{Data Curation and Quality Filtering}\label{sec::data_curation}
High data quality is critical for training a world model that produces temporally consistent, artifact-free rollouts. We apply a multi-stage filtering pipeline to remove problematic sequences before training. We discard any sequences with unreliable sensor data (e.g., jumps in the ego trajectory), unreliable annotations (e.g., high uncertainty in auto-labeled objects), or prediction disagreements. Next, we use a VLM to detect undesirable visual artifacts in the camera data, such as chromatic aberration, and remove affected clips. As a final step, we perform data de-duplication based on ego-trajectory and visual features to down-sample repetitive clips in the dataset, such as straight highway driving.

\subsection{Dataset Composition}\label{sec::data_composition}
\modelname training dataset spans 15 countries across North America, Europe, and Asia (see \cref{tab:data_stats}).
RDS provides the mid-training data (16,600\,h; 3M clips). Post-training and finetuning use RDS-HQ-1M (4,944\,h; 1,142,285 clips), of which 504,488 are 10\,s clips ($\approx$44\%) and 637,797 are 20\,s clips ($\approx$56\%).
We hold out 5,000 clips from the training set for evaluation and testing. Rather than sampling proportionally from the training distribution, these clips are balanced across various scenario categories covering road users (cyclists, pedestrians, motorcycles, strollers), large vehicles (trucks, trailers), weather and lighting (rain, snow, fog, night), and rare infrastructure (tunnels, railroad crossings, construction zones, accident scenes). For 300 of those clips, we additionally sourced a longer (60\,s) version, which are used for long-term temporal consistency evaluation.
\cref{tab:data_stats} summarizes the final training dataset.

\begin{table}[ht]
\centering
\caption{Summary of the \modelname training dataset.}
\label{tab:data_stats}
\begin{tabular}{l r}
\toprule
\textbf{Statistic} & \textbf{Value} \\
\midrule
Total driving hours (RDS)         & 16,600 \\
Total driving hours (RDS-HQ-1M)         & 4,944 \\
Number of sequences         & 504,488 (10s) + 637,797 (20s) = 1,142,285  \\
Frame resolution            & $704 \times 1280$ \\
Camera views per frame      & 7 in total, 4 in training \\
Geographic regions          & 15 countries: US, DE, JP, KR, GB, FR, ES, SE, PT, DK, FI, PL, IT, AT, BE \\
\bottomrule
\end{tabular}
\end{table}

\subsection{Curation and Inspection with SIL-Wheel}\label{sec::silwheel}
Building an \modelname training mixture and the matching held-out evaluation sets requires reasoning about millions of candidate clips along multiple axes: geography, weather, time of day, ego behavior, agent density, and rare-but-safety-critical scenarios such as collisions or vulnerable road users. We use NVIDIA SIL's video search and curation platform, SIL-Wheel~\citep{nvidia2026silwheel}, as the workbench for this selection. SIL-Wheel indexes the underlying corpora (including RDS and RDS-HQ-1M) at clip granularity and exposes a set of composable search primitives, spanning caption search, dense semantic and visual retrieval, ego-trajectory queries, classifier scoring, and perception-based filters, that can be combined within a single query.

For \modelname, SIL-Wheel plays three concrete roles:
\begin{itemize}
    \item \textbf{Finetuning slice construction.} We compose post-training mixtures by retrieving clips that match conditions that the natural-data distribution under-represents, including rare weather, construction zones, vulnerable road users, articulated motion, and complex multi-agent interactions, and upweighting those slices in the post-training corpus.
    \item \textbf{Eval-set slice construction.} The held-out evaluation set described in \cref{sec::data_composition} is decomposed into slice-aware buckets through the same search interface. This enables the long-tail and per-scenario reporting in \cref{sec::results}, alongside aggregate metrics.
    \item \textbf{Data inspection.} The clip-centric UI is used to validate autolabel quality on RDS-HQ-1M, triage clips surfaced by the quality-filtering pipeline of \cref{sec::data_curation}, and visually verify that targeted slices contain what we intended before they enter the training mixture.
\end{itemize}

\section{Model Architecture}
\label{sec::model}

\begin{figure}[t]
    \centering
    \includegraphics[width=\linewidth]{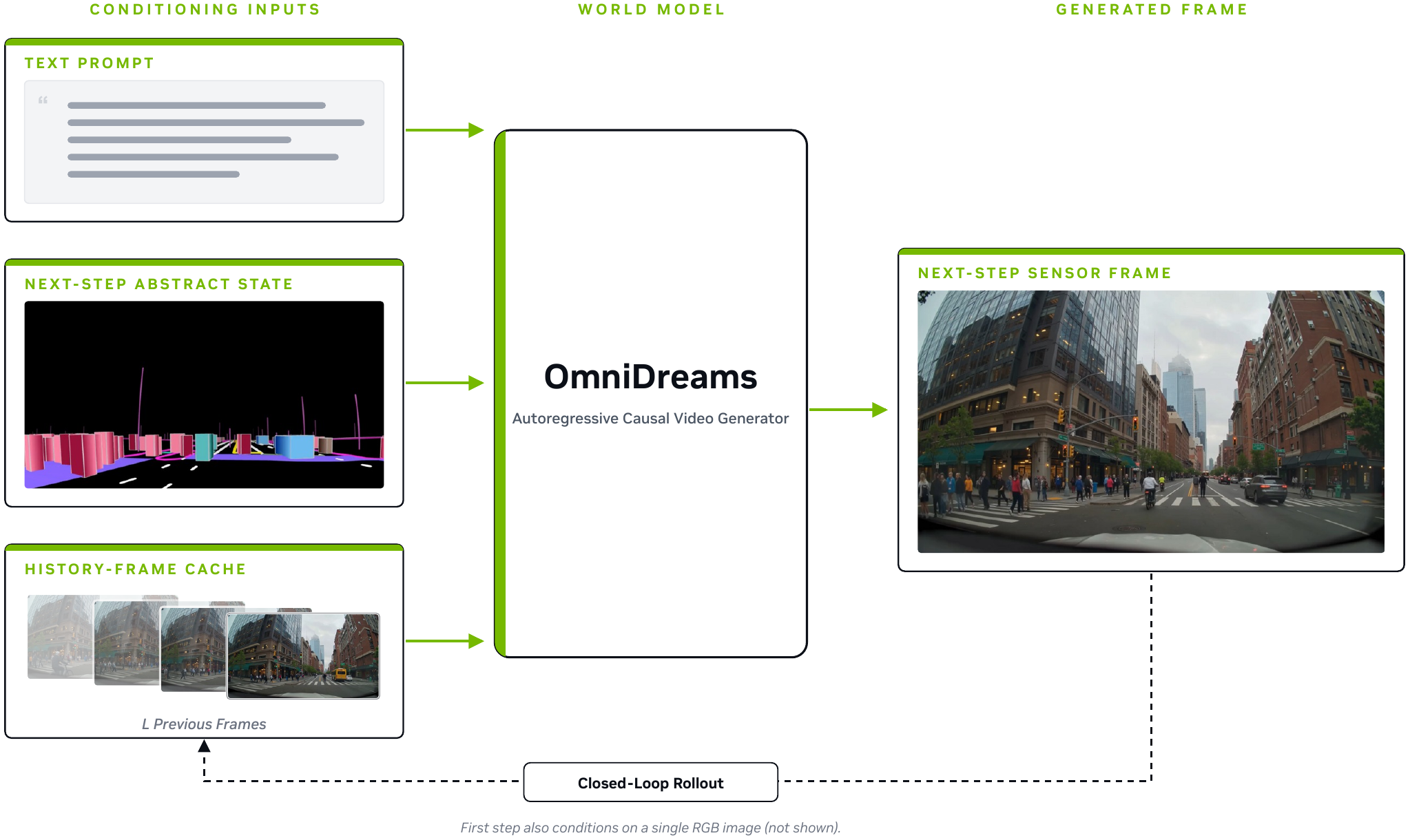}
    \caption{\modelname conditions on three inputs: a text prompt, the next abstract state from the simulator, and a history-frame cache to generate next-step sensor frames that are returned to the policy in a closed loop. In the first generation step, \modelname also conditions on a single RGB image, which is not shown in this figure.}
    \label{fig:input_output}
\end{figure}

A key requirement for closed-loop simulation is interactivity, allowing the model to respond immediately to changes in actions from the driving policy model or human input. To support this, \modelname performs generation in an autoregressive manner. At each time step, the simulator updates the world state based on the latest action, and the model generates a short sequence of future frames conditioned on this updated state.

This design differs from offline video generation methods, such as~\citep{openai2024sora, nvidia2025worldsimulationvideofoundation, deepmind2025veo3, wan2025wan, polyak2024movie}, which produce long clips using bidirectional and diffusion-based sampling. Instead, \modelname adopts a causal diffusion formulation, in which each prediction depends only on past observations and the current conditioning. Temporal consistency is achieved through a streaming KV cache. By reusing previously computed attention keys and values, the model maintains long-term context without recomputing the full sequence, enabling efficient and stable rollouts over extended horizons.

We provide two variants of \modelname:
\textbf{(1) Single-view model (\modelnamesv)}: Generates a single front-facing camera view, producing 8 frames per step (2 latent frames);
\textbf{(2) Multi-view model (\modelnamemv)}: Jointly generates four synchronized views (front camera, cross-left camera, cross-right camera, front-telescope camera), producing 16 frames per step (4 latent frames).

\subsection{Network Architecture}

\modelname conditions on the following inputs, shown in \cref{fig:input_output}:

\begin{itemize}
\item \textbf{First-frame RGB}: The first frame of the simulation session, encoded as clean latent tokens to initialize the generation
\item \textbf{Text prompt}: A high-level description of environmental attributes such as lighting, weather, and time of day. The text is encoded using the Cosmos text encoder and injected via cross-attention layers
\item \textbf{Abstract world scenario}: A structured representation of the simulator state that is aligned to the video being generated by \modelname, including maps (e.g., lane lines, traffic poles) and dynamic agents represented as bounding boxes (see \cref{fig:teaser}). The static maps and dynamic agent trajectories are obtained in advance (such as via real recordings from a vehicle), and the driving action produced by the policy model or human driver is used to render the world-scenario conditioning video input
\item \textbf{Memory cache}: A streaming KV cache storing previously generated tokens for temporal context
\end{itemize}

The first-frame latent and conditioning signals are combined with noisy latent tokens, then processed by a causal transformer backbone similar to that of the Cosmos-Predict 2.5 base model~\citep{nvidia2025worldsimulationvideofoundation}.

\modelname incorporates a lightweight control branch to efficiently inject structured world conditioning into the generative model. Instead of processing control inputs with a separate network (e.g., ControlNet), the structured simulator state is first encoded using a small MLP into compact control tokens. These tokens are aligned with the latent representation and concatenated with the visual tokens before being fed into the transformer. This design introduces minimal computational overhead while enabling effective conditioning on scene structure. It also stabilizes autoregressive generation by maintaining a clear separation between control signals and visual content. In practice, the lightweight control branch allows efficient integration of the simulator state while preserving the throughput required for real-time generation.

\subsection{Consistent Multi-View Generation}

\modelname supports joint generation of multiple camera views while maintaining cross-view consistency and real-time performance.
\label{learnable_view_emb}
Each camera view is associated with a learnable embedding that is added to the token representation to indicate the view identity. In addition, each view has its own conditioning inputs, including a text prompt, the first-frame RGB image, and a structured world state corresponding to its camera configuration. This allows the model to represent both shared scene structure and view-specific appearance.

A naive multi-view formulation applies full self-attention across all tokens over time and views, resulting in quadratic complexity $\mathcal{O}(N^2 T^2)$ with respect to the number of views $N$ and temporal length $T$. This quickly becomes impractical for real-time simulation.
To address this, \modelname factorizes attention into two components:

\textit{Temporal attention} is applied independently within each view. Tokens attend to previous frames through the causal KV cache, capturing motion dynamics and maintaining temporal consistency.

\begin{figure}[t]
    \centering
    \includegraphics[width=0.98\linewidth]{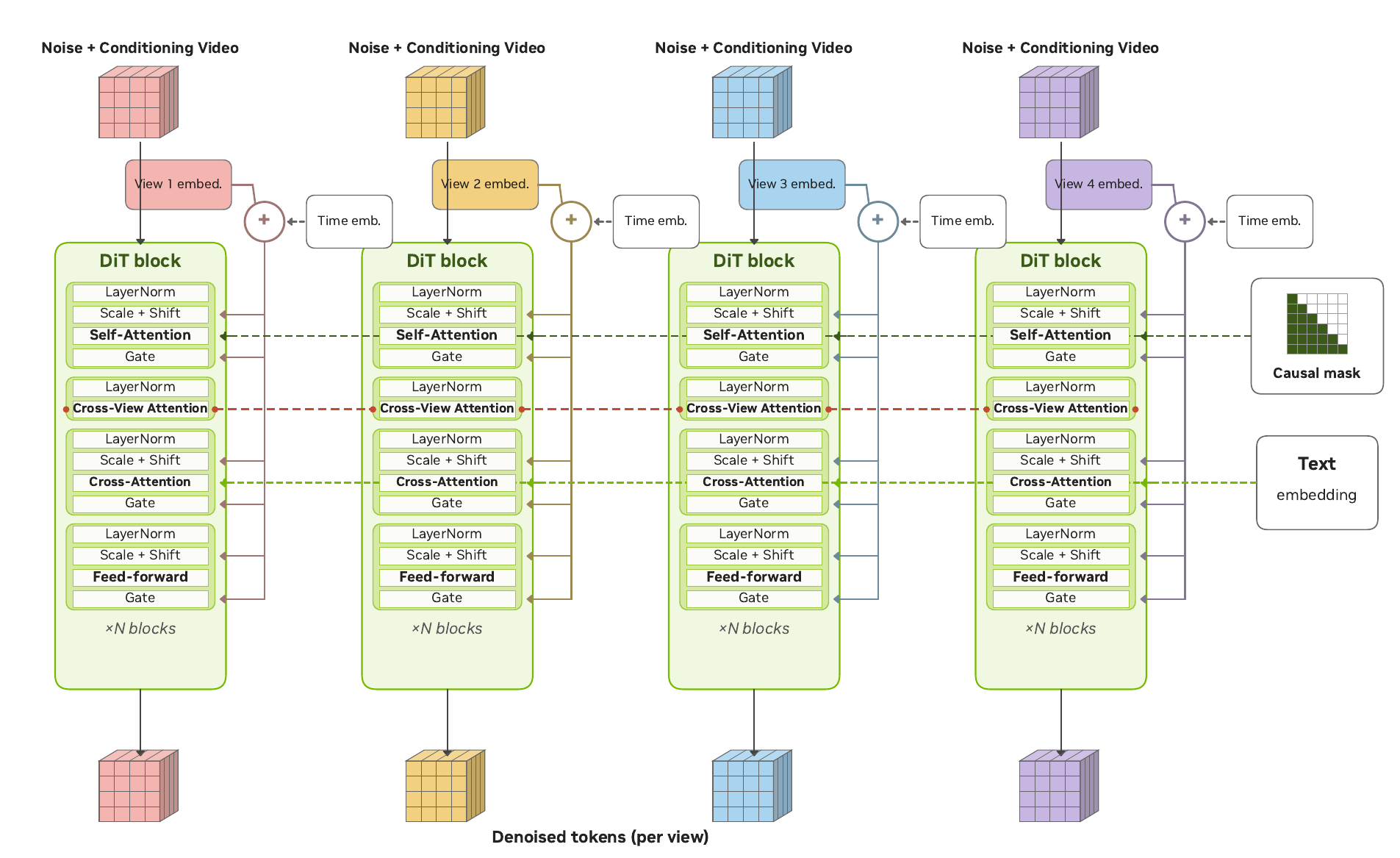}
    \caption{Multi-view \modelname DiT. Compared to the single-view Cosmos-Predict 2.5 backbone, each Multi-View Cross Block adds (i) a per-view view embedding that is summed with the time embedding and used as an additive AdaLN signal on the shift, scale, and gate of all sublayers, and (ii) a Cross-View Attention sublayer after the text Cross-Attention that attends across the tokens of all V views to enforce cross-camera consistency.
}
    \label{fig:multiview}
\end{figure}

\textit{Cross-view attention} is applied across views at each time step. Tokens from different views attend to one another, enabling a consistent representation of shared scene elements such as geometry, object positions, and motion. \cref{fig:multiview} gives a summary of our \modelnamemv.

This factorization reduces the overall complexity to: \[ \mathcal{O}(N T^2) + \mathcal{O}(N^2), \] significantly improving efficiency compared to full joint attention.

This design enables scalable multi-view generation with strong cross-view consistency, while supporting higher resolution, longer temporal context, and more views. In practice, \modelname supports up to 7 synchronized views for surround-view simulation.

\section{Training}
\label{sec::training}
We employ a multi-stage training strategy to efficiently leverage diverse data sources in building \modelnamesv and \modelnamemv. From the bidirectional video generation model, Cosmos-Predict 2.5, we first mid-train it on the RDS dataset to improve its AV capability, and then add another mid-training stage that adapts the model for bidirectional multi-view generation by incorporating a cross-view attention architecture (for \modelnamemv only). This model is then adapted for autoregressive generation via a Diffusion Forcing mid-training stage. Controllable generation via world-scenario map control is then added in post-training stages to both the bidirectional and autoregressive models. Lastly, a distillation stage with Self Forcing DMD enables few-step generation while removing autoregressive error accumulation. After our training, we can roll out minutes-long videos without significantly sacrificing quality, shown in \cref{fig:ar} at inference time.

\subsection{World-Scenario Control and Multi-view adaptation}

\paragraph{Preamble}
Following \citep{nvidia2025worldsimulationvideofoundation}, \modelname-MV uses the rectified-flow~\citep{liu2022flow} objective to train the model. Formally, let $\mathrm{x}$ denote a latent video, $\epsilon \sim \mathcal{N}(0, I)$ a noise sampled from the standard normal distribution, and $t \in [0, 1]$ a time-step drawn from a logit-normal distribution. The flow matching objective is $L = \mathbb{E}_{\mathrm{x}, t} [ \|\mathbf{u}{\theta}(\mathrm{x}_t, t) - \mathrm{v}_t\|^2 ]$, where $\mathrm{x}_t = (1-t)\mathrm{x} + t\epsilon$ and $\mathrm{v}_t = \epsilon - \mathrm{x}$. The model can also be conditioned on auxiliary information $\mathrm{c}$, denoted as $\mathbf{u}{\theta}(\mathrm{x}_t, t; \mathrm{c})$, which includes text caption, first-frame images, and control signals.

\paragraph{Multi-view adaptation}

\modelname-MV supports generating videos from front-wide, cross-left, cross-right, and front-telescope camera sensors. These videos (particularly the cross-left, cross-right, and front-telescope) have unique intrinsics and motion that are rarely encountered in general-domain videos. Hence, a mid-training stage is needed to bestow the model with knowledge of multi-view videos. Starting from Cosmos-Predict 2.5, we enable learnable view embeddings (\cref{fig:multiview}) and train the model on a uniform mix of front-wide, cross-left, cross-right, and front-telescope clips. This is followed by adding cross-view attention layers to the network and training on multi-view videos, during which the cross-view attention layers learn cross-view correspondences and consistency.

View embedding is injected into the tokens through adaptive layer normalization in each transformer block, similar to how timestep embedding is injected. These view embeddings are zero-initialized to ensure stable convergence when they are added to the network. Cross-view attention layers are added to each block after the cross-attention layer and before the MLPs. The output projection weights of each cross-view attention layer are also zero-initialized for stable convergence. We train on a 1:1 mixture of text-to-video and image-to-video tasks to support first-frame conditioning while preserving the model's capacity to generate novel content.

\paragraph{World-scenario map control}
A world-scenario map is needed in both \modelnamesv and \modelnamemv for the video generation models to function as world simulation models. Using pre-trained or multi-view-adapted weights, we append a world-scenario control branch initialized to zero and train with the flow-matching objective. For training efficiency, we first train on 93-frame clips until convergence, then expand to 189-frame clips to learn longer-term temporal consistency. The result of this training is World-Scenario controlled bidirectional models, which are suitable as teacher models for further distillation.

\subsection{Mid-training for Autoregressive Generation}
\begin{figure}[t]
    \centering
    \includegraphics[width=0.95\linewidth]{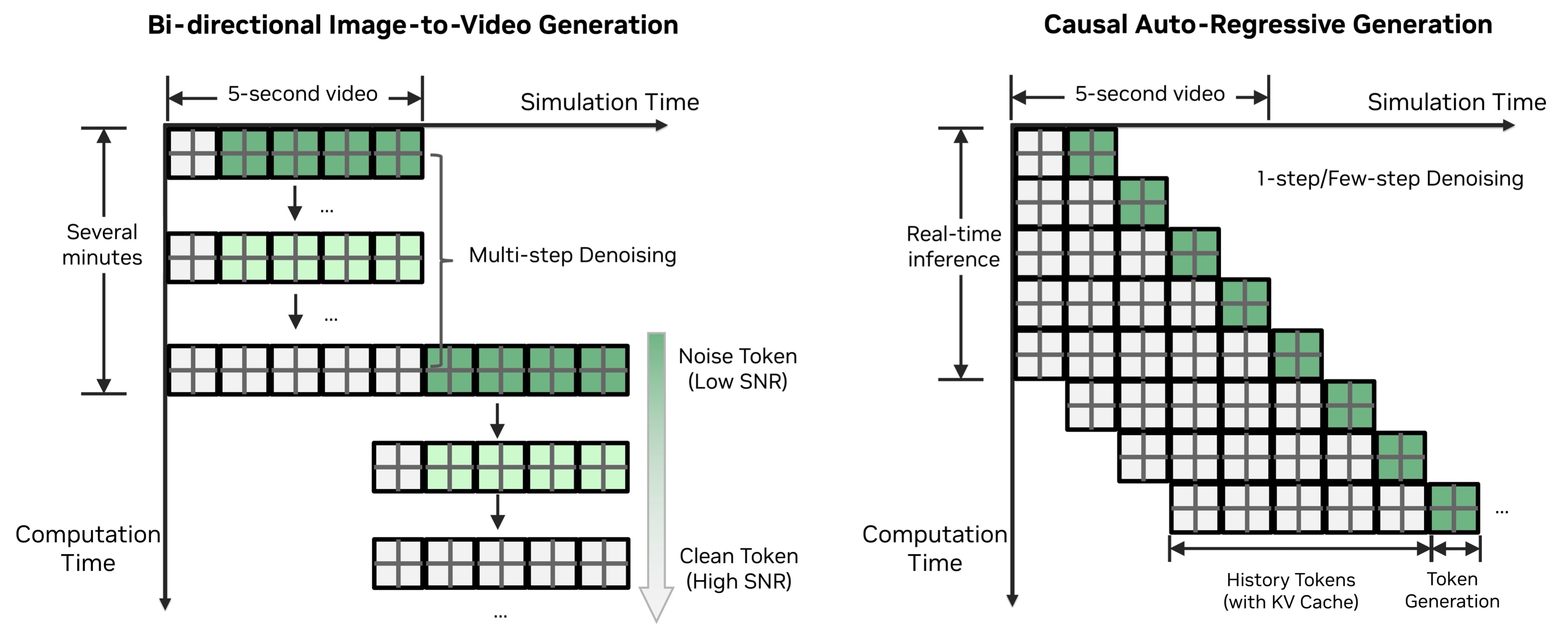}
    \caption{Autoregressive video generation diagram (right) comparing bidirectional image-to-video denoising (left) and causal KV-cache based generation for consistent long video rollouts.}
    \label{fig:ar}
\end{figure}
We apply causal masking with Diffusion Forcing~\citep{chen2024diffusionforcing} training to convert the bidirectional models into causal models for autoregressive generation. Empirically, we found that this training benefited from extensive optimization beyond the scale of the available World-Scenario labeled data. Hence, we perform this training from model weights without world-scenario control on the RDS dataset, and then add the control branch and continue Diffusion Forcing training using the RDS-HQ-1M dataset.

\paragraph{Causal Masking}
Let $\mathrm{x}^{1:T}$ denote a latent video of $T$ latent frames. In autoregressive video generation, we factorize the distribution of full video sequences as $p(\mathrm{x}^{1:T}) = \Pi_{i=1}^T p(\mathrm{x}^{i}|\mathrm{x}^{<i})$, and parameterize the conditional factor $p(\mathrm{x}^{i}|\mathrm{x}^{<i})$ as a flow matching model $\mathbf{u}\theta(\mathrm{x}_t^{i}|\mathrm{x}^{<i})$ initialized from a bidirectional video generation model. Causal masking is applied within the network's self-attention layers to ensure that tokens of $\mathrm{x}^i$ can only attend to tokens of frames $\leq i$, shown in \cref{fig:ar}. This formulation can be generalized to block-autoregressive generation, in which blocks of $k$ frames are generated together, i.e. $p(\mathrm{x}^{1:T}) = \Pi_{j\in\{ki\}, i\in 1:T} p(\mathrm{x}^{j:j+k}|\mathrm{x}^{<j})$.  In practice, we implement causal masking with Flex-Attention~\citep{dong2024flex}.

\paragraph{Diffusion Forcing}
We sample noise $\epsilon$ as in flow matching, and draw a vector of diffusion times $\mathrm{t} = [t_i],\  i \in 1:T $ independently from a log-normal distribution. Let $\mathrm{x}^{1:T}_\mathrm{t} = (I-\mathrm{t}) \cdot \mathrm{x}^{1:T} + \mathrm{t} \cdot \epsilon$, and $\mathrm{v}_\mathrm{t} = \epsilon - \mathrm{x}^{1:T}$, then the Diffusion Forcing objective can be written as:
\begin{equation}
    \mathbf{L}_{DF} = \mathbb{E}_{\mathrm{x}^{1:T}, \epsilon} \left[ \|\mathbf{u}\theta(\mathrm{x}^{1:T}_\mathrm{t}, \mathrm{t} ) -  \mathrm{v}_\mathrm{t}\|^2\right].
\end{equation}
Efficient computation of $\mathbf{u}\theta(\mathrm{x}^{1:T}_\mathrm{t}, \mathrm{t} )$ in one forward pass is achieved through causal masking, in which the computation of $\mathbf{u}\theta(\mathrm{x}^{i}_\mathrm{t}, \mathrm{t}|\mathrm{x}^{<i}_\mathrm{t})$ is reused in parallel by $\mathbf{u}\theta(\mathrm{x}^{j}_\mathrm{t}, \mathrm{t}|\mathrm{x}^{<j}_\mathrm{t}),\ j>i$.

\subsection{Distillation}
\label{sec::distillation}
To enable few-step generation while eliminating the compounding errors that arise in long autoregressive rollouts, we apply Self Forcing~\citep{huang2025self}, a training framework that combines autoregressive self-rollout with a holistic, video-level distribution-matching objective based on Distribution Matching Distillation (DMD)~\citep{yin2024one}.

\paragraph{Self Forcing Training via Self-Rollout.}
Standard teacher forcing trains diffusion models by conditioning each frame's denoising on clean, ground-truth context frames. At inference time, however, the model must condition on its own previously generated (and thus imperfect) outputs, creating a distribution mismatch that causes compounding errors over long rollouts---a phenomenon known as \emph{exposure bias}~\citep{huang2025self}. Self Forcing~\citep{huang2025self} closes this train-test gap by replacing ground-truth context with self-generated frames during training: the model performs an autoregressive self-rollout, generating each frame conditioned on its own previously generated outputs via a $K$-step diffusion process ($K{=}2$, with a timestep schedule of $[1000, 450]$ in our implementation). To keep training tractable, gradient backpropagation is restricted to a single randomly sampled denoising step $s \sim \mathrm{Uniform}(1,\ldots,T)$ per training iteration, ensuring that all intermediate steps receive supervision across iterations. Gradients are additionally detached from the KV cache embeddings of previous frames, confining gradient flow to the current frame. Long-video generation is supported via a rolling KV cache~\citep{huang2025self}: we maintain a fixed-size cache of the most recent $L$ frames, evicting the oldest entry whenever a new frame is added. This reduces inference complexity from $\mathcal{O}(TL^2)$ to $\mathcal{O}(TL)$ and enables generation of arbitrarily long sequences. To match inference behavior during training, we restrict attention to the same rolling window.

\paragraph{Holistic Distribution Matching via DMD.}
Rather than supervising individual denoised frames with a pixel-wise reconstruction loss, Self Forcing employs a holistic, video-level distribution matching objective. Concretely, it minimizes the reverse KL divergence between the distribution of self-rolled-out video clips $p_\theta$ and the real data distribution $p_\mathrm{data}$, using the DMD formulation~\citep{yin2024one}:
\begin{equation}
    \mathcal{L}_\mathrm{DMD}(\theta) = \E\!\left[\tfrac{1}{2}\left\|\hat{x} - \mathrm{sg}\!\left[\hat{x} - \bigl(\mathbf{f}_\psi(\hat{x}_t, t) - \mathbf{f}_\phi(\hat{x}_t, t)\bigr)\right]\right\|^2\right],
    \label{eq:dmd}
\end{equation}
where $\hat{x}$ is the self-rolled-out video clip, $\mathbf{f}_\phi$ is a frozen real score network, $\mathbf{f}_\psi$ is a learned fake score network trained to estimate the score of the generated distribution, and $\mathrm{sg}[\cdot]$ denotes the stop-gradient operation. The score difference $\mathbf{f}_\psi - \mathbf{f}_\phi$ provides a gradient that steers the generator toward the real data manifold at the full-sequence level, without requiring paired data or pixel-wise supervision.

\paragraph{Progressive Training with a Longer Teacher.}
While the Self Forcing distillation procedure above substantially reduces exposure bias for short rollouts, we observe that the resulting model still exhibits \emph{shifting artifacts}---temporal inconsistencies that accumulate when the rolling KV cache extends beyond the training context window during long generation. To mitigate this, we adopt a progressive teacher strategy. We first train a bidirectional (non-causal) video model with a substantially longer temporal context window; as a bidirectional model, it does not suffer from rolling-cache artifacts. We then use this long-context bidirectional model as the teacher to continue finetuning the model that was previously distilled with the short-context teacher. As shown in \cref{fig:long_rollout}, this progressive strategy significantly reduces shifting artifacts in long autoregressive rollouts.

For the distillation stage, we use a high-quality and challenging subset from our full training corpus with $58k$ videos. Concentrating distillation on this subset allows \modelname to better adapt to the perceptually demanding conditions of complex urban environments such as dense traffic, varied lighting, and intricate road geometry while retaining generation quality on highway and rural scenarios.

\section{Training-free Model Inference Optimization}
\label{sec::inference}

This section describes the inference system that runs the few-step distilled Cosmos-Predict 2.5 backbone (\cref{sec::training}) in our closed-loop simulator. Everything below is \emph{training-free}: no diffusion-model weights are touched relative to the distilled checkpoint. The deployed system generates a 16-frame four-view chunk at 704$\times$1280 in 151\,ms on a 16-GPU NVIDIA GB300 cabinet (\textbf{105 effective FPS per camera}) and an 8-frame single-view chunk in 118\,ms on a single GB300 (\textbf{68 effective FPS}). 

\subsection{Model Optimization}
\label{sec::inf_model}

\paragraph{Local temporal attention.}
\label{sec::inf_local_attn}
Full temporal attention over an unbounded rollout is impractical for both compute and cache size. We use local-window attention with a fixed window of 6 latent frames (24 RGB frames) for \modelnamesv and 8 latent frames (32 RGB frames) for \modelnamemv. The window size is a tradeoff between memory and speed.

\paragraph{Streaming static-shape KV cache.}
\label{sec::inf_kvcache}
The DiT is autoregressive: each chunk reuses keys/values from previous chunks. We pre-allocate the cache to the fixed size and keep its tensor shape static across the whole rollout, and move cache updates on a separate thread to allow the main thread free from cache update latency.

\paragraph{Compile and CUDA Graphs.}
\label{sec::inf_compile}
Because the KV cache is pre-allocated to a fixed local-attention window (\cref{sec::inf_kvcache,sec::inf_local_attn}), the DiT forward runs on tensors of a single static shape for the entire rollout. We compile it with \texttt{torch.compile}~\citep{ansel2024pytorch2} and capture it into a CUDA Graph~\citep{nvidia_cuda_graphs}; the graph is captured lazily on the first chunk and reused for every subsequent chunk and every rollout of the same shape.

\paragraph{Lightweight encoders and decoders.}
\label{sec::inf_scenario}
\label{sec::inf_tae}
We swap the original VAE used in Cosmos with LightVAE or LightTAE from LightX2V~\citep{lightx2v}. LightVAE encodes the conditioning video (the initial frame and the world-scenario condition videos, \cref{sec::data}) into latent space for \modelnamesv and offers the best balance between reconstruction quality and speed. \modelnamemv uses pixel shuffle technique to replace LightVAE to further speed up the model. Its latency is negligible.
We use LightTAE to decode the denoised latents back to RGB for both \modelnamemv and \modelnamesv and is the best for its extremely low decode latency.

\paragraph{Hoisting step-invariant operators.}
We hoist step-invariant computations out of the denoising loop. Self-attention at every denoising step uses the same RoPE frequencies, so we compute them once per chunk instead of once per step. Likewise, the patchify and unpatchify operators that convert latents into and out of the DiT's token layout depend only on the chunk shape, so we run them once at the start and end of chunk generation rather than on every entry and exit of the DiT.

\subsection{Multi-GPU Inference}
\label{sec::inf_mgpu}

The DiT exposes three natural parallelism axes: the camera-view axis $V$ (at most the number of cameras), the temporal axis $T$ within a chunk (at most the number of latent frames per chunk), and the spatial axis $HW$ inside one frame's attention. We use a hierarchical context-parallel strategy that shards along all three. Our 16-GPU four-view configuration is $V{=}4$, $T{=}4$, $HW{=}1$.

Ranks are added in the order $V \to T \to HW$, chosen by how each axis interacts with attention cost. Self-attention dominates the DiT and is per-camera, so sharding along $V$ never splits a self-attention call: 4 ranks along $V$ give close-to-linear speed-up in our 4-camera setting and saturate this axis at 4 GPUs. Beyond that, the remaining work sits inside one camera, leaving $T$ or $HW$. We prefer $T$ because temporal sharding leaves the per-time-step cross-view attention block intact, whereas $HW$ sharding splits both self- and cross-view attention. With 4 latent frames per chunk in four-view inference, $V{\times}T = 4{\times}4 = 16$ already saturates 16 GPUs with $HW{=}1$, and we would only start sharding $HW$ beyond 16 GPUs. For the context-parallel attention itself we use an in-house ring-attention implementation~\citep{liu2024ringattention}\label{sec::inf_ring} that overlaps KV-shard transfer with local attention compute.

\subsection{End-to-End Performance}
\label{sec::inf_results}

\Cref{tab::inf_sv_scaling,tab::inf_4view_scaling} report per-chunk latency on NVIDIA GB300. Both runs use the 2-step diffusion model from \cref{sec::training} at 704$\times$1280: \modelnamemv uses 16 RGB frames per chunk with a local-attention window of 8 latent frames, and \modelnamesv uses 8 frames per chunk with a window of 6 latent frames. Single-view inference already achieves the real-time performance (defined as 30 FPS) on a single GPU, so multi-GPU is only needed for \modelnamemv, where the 16-GPU configuration reaches 105 effective FPS per camera. We report single-view model's multi-GPU scaling behavior nevertheless.

\begin{table}[t]
  \centering
  \small
    \begin{tabular}{lcccc}
      \toprule
      Stage & 1$\times$GPU & 2$\times$GPU & 4$\times$GPU & 8$\times$GPU \\
      \midrule
      World scenario encoding              &  28\,ms &  26\,ms &  26\,ms &  26\,ms \\
      Diffusion DiT                        &  84\,ms &  71\,ms &  49\,ms &  47\,ms \\
      RGB Decoder                          &   6\,ms &   5\,ms &   5\,ms &   5\,ms \\
      KV-cache update (separate thread)    &  42\,ms &  34\,ms &  23\,ms &  22\,ms \\
      \midrule
      Total                                & 118\,ms & 102\,ms &  80\,ms &  78\,ms \\
      Effective FPS                        &      68 &      78 &     100 &     103 \\
      \bottomrule
    \end{tabular}
  \caption{Per-chunk (8 RGB frames / 2 latent frames) timings for single-view inference on NVIDIA GB300. KV-cache update is excluded from ``Total'' because it runs off the hot path. ``Effective FPS'' is $K / \text{Total}_\text{ms}$ where $K{=}8$ is the chunk size.}
  \label{tab::inf_sv_scaling}
\end{table}

\begin{table}[t]
  \centering
  \small
  \begin{tabular}{lcccc}
    \toprule
    Stage & 1$\times$GPU & 4$\times$GPU & 8$\times$GPU & 16$\times$GPU \\
    \midrule
    Diffusion DiT      & 1,184\,ms & 300\,ms & 179\,ms & 121\,ms \\
    RGB Decoder        &  105\,ms &  30\,ms &  30\,ms &  30\,ms \\
    KV-cache update (separate thread)    &  558\,ms & 149\,ms &  91\,ms &  67\,ms \\
    \midrule
    Total              & 1,289\,ms & 330\,ms & 209\,ms & 151\,ms \\
    Effective FPS      &       12 &      48 &      74 &     105 \\
    \bottomrule
  \end{tabular}
  \caption{Per-chunk timings (16 RGB frames / 4 latent frames)  for four-view inference on NVIDIA GB300. KV-cache update is excluded from ``Total'' because it runs off the hot path. ``Effective FPS'' is $K / \text{Total}_\text{ms}$ where $K{=}16$ is the chunk size, expressed per camera. In \modelnamemv, we used pixel shuffle technique to encode conditioning video, and its latency is negligible ($<1 $ms)}
  \label{tab::inf_4view_scaling}
\end{table}

\subsection{FlashDreams: Generalized Inference and Serving Infra}
\label{sec::inf_flashdreams}

The recipes above, including streaming static-shape KV cache, local-window attention, CUDA-graph capture, is not specific to our model architecture. We have packaged it into \emph{FlashDreams}~\citep{flashdreams}, an open-source streaming inference stack for autoregressive world and video models, and validated it on additional Wan2.1-based backbones~\citep{wan2025wan}. On Self Forcing~\citep{huang2025self}, a Wan2.1-based autoregressive text-to-video model, FlashDreams reaches up to $1.95\times$ speed-up over the official implementation on a single GB200. On Lingbot-World~\citep{lingbot-world}, a Wan2.1-14B-based autoregressive camera-control world model, it reaches up to $2.49\times$ speed-up on 4 H100s. Beyond the per-chunk model optimizations, FlashDreams also ships the end-to-end serving stack described in \cref{sec::applications}: a gRPC and webRTC server/client protocol that streams control inputs from the client and RGB frames back from the server. We hope FlashDreams serves as a practical, extensible starting point for the community to deploy autoregressive world and video models in latency-critical settings, and we welcome contributions to grow the set of supported backbones.

\section{Closed-Loop Simulation Integration} 
\label{sec::applications}

\begin{figure}[t]
  \centering
  \includegraphics[width=0.98\linewidth]{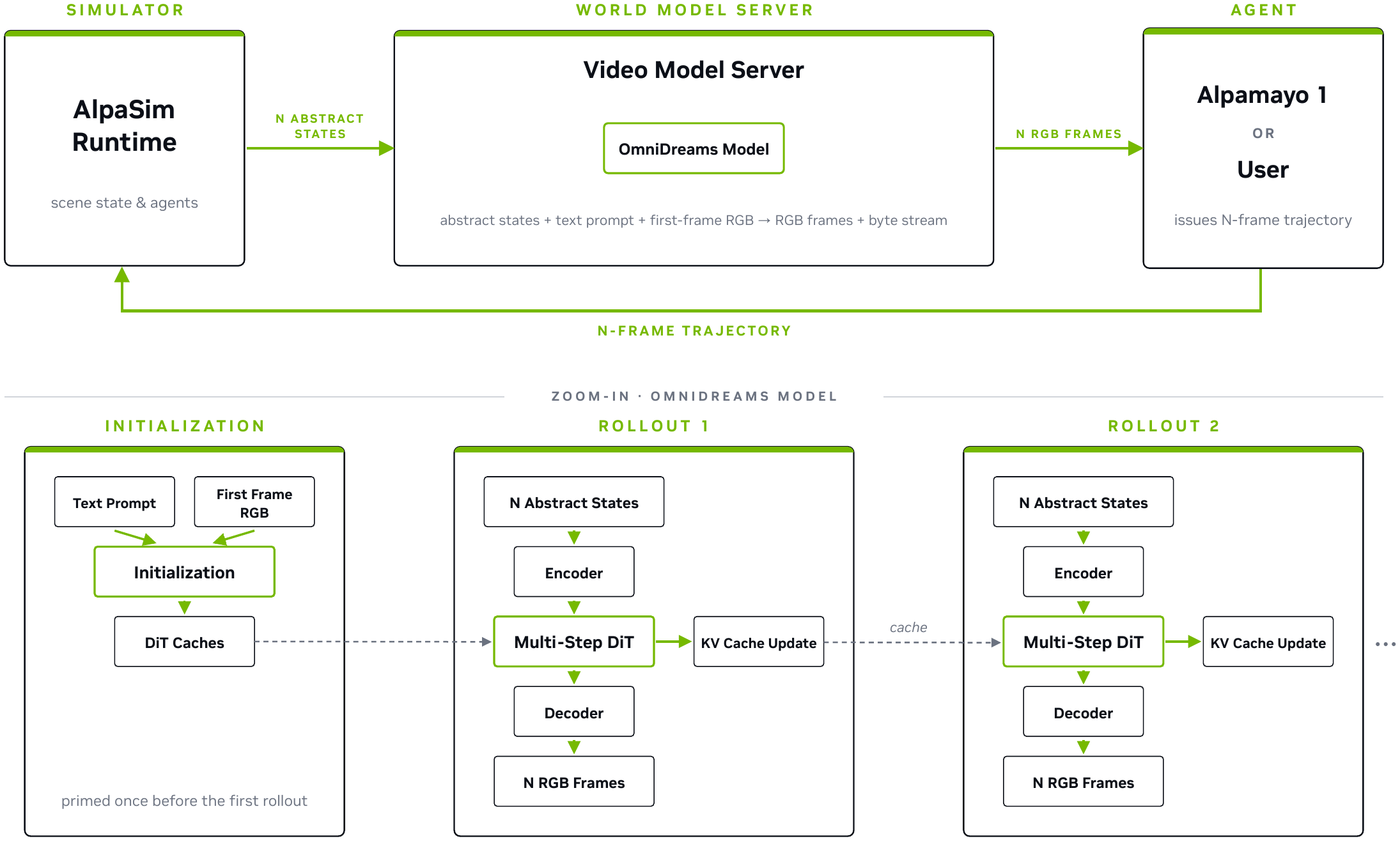}
  \caption{End-to-end inference pipeline over two consecutive client/server round-trips. KV-cache maintenance runs off the critical path on a side thread. A chunk is $K$ contiguous frames generated together; we use $K{=}16$ for \modelnamemv (all cameras generated jointly) and $K{=}8$ for \modelnamesv.}
  \label{fig::inf_pipeline}
\end{figure}

For closed-loop use, the video model is packaged into a stateful \emph{video model server} that talks to a simulator client over gRPC. \Cref{fig::inf_pipeline} shows two consecutive round-trips. In each round-trip, the client sends a trajectory from the policy model, a text prompt, and an initial frame (only on the very first chunk); the server renders the world-scenario conditioning, runs two denoising steps of a causal DiT with a streaming KV cache, decodes the latents to RGB, and returns JPEG-encoded frames. The client then advances its world state, queries its driving policy for the next trajectory, and sends the next request. The server is stateful for the whole rollout: the KV cache, captured CUDA Graphs, and renderer state are all reused chunk-to-chunk, so each follow-up request carries only the new trajectory and prompt.

This poses unique challenges, stemming from three main properties of video models as renderers:
\begin{itemize}
\item Distributed inference --- for high frame rates, video models are run with multiple processes (typically one per GPU) and use networking for inter-process communication. This precludes the standard in-process integration of video models (e.g.\ as a Python dependency), as that would require the simulation layer to become distributed as well.
\item Autoregressive nature --- video models are naturally internally stateful, preventing rendering frames in an arbitrary order. This requires the simulator to keep track of that state and avoid invalidation.
\item Chunk-based generation --- for performance and quality reasons, the current generation of video models generates frames in \textit{chunks}, which correspond to underlying VAE temporal compression. This implies quantizing time into intervals over which the agent's behavior must be fixed and known at the start of each interval.
\end{itemize}

In the sections below, we describe how we approach each of these issues in our prototype integration with NVIDIA AlpaSim~\citep{nvlabs2025alpasim}, a research-oriented AV simulator.

\subsection{AlpaSim}

\href{https://github.com/NVlabs/alpasim}{NVIDIA AlpaSim} is an open-source, research-oriented AV simulator. To maximize software modularity and deployment flexibility, AlpaSim adopts a microservice architecture: its core runtime is lightweight and uses remote procedure calls (RPCs) to specialized services, such as camera frame renderers, AV policy, and physics simulation, via the gRPC protocol. The entire system is deployed as a set of Docker containers, allowing maximum flexibility in Python and OS environments, as well as deployment (including hosting different components in different data centers).

AlpaSim is built with throughput in mind. Thanks to its use of asynchronous RPC requests, launching multiple concurrent rollouts allows Python's event loop to dynamically serialize them across microservices, automatically implementing pipeline parallelism. This imposes a requirement on individual services to keep track of their state across multiple, potentially out-of-order, rollout requests.

Our integration of \modelname involves replacing NVIDIA NuRec as the camera frame renderer, defining new gRPC protocols to manage the rendering state, and making AlpaSim's runtime loop more flexible to accommodate the chunk semantics of video models.

\subsection{Networked integration in AlpaSim}
The standard approach to running multi-GPU deployments spanning multiple nodes is rank-based parallelism: the same Python script is executed in parallel (usually once per GPU), with environment variables indicating its rank, the total number of processes, and the address of the master process. The script contains conditional statements which influence its behavior, e.g.\ restricting each rank to processing a subrange of all video tokens. This pattern does not naturally extend to using the video model as a subcomponent in a larger system, e.g.\ a simulator which contains state-keeping logic to be executed only once and other submodules (such as the AV policy or traffic simulation) which may want to launch with their own rank-parallel logic.

Our solution is to add a second layer to the networking hierarchy via gRPC. We implement a gRPC server running on rank 0 of the video model deployment that receives rendering requests and forwards them to the remaining ranks via NCCL events. Once rendering a chunk is complete, the generated frames are gathered on rank 0, serialized, and sent back to AlpaSim. This solution avoids unifying the dependencies, code, and rank dispatch logic of \modelname with the remaining components of AlpaSim.

Future work will improve this scheme by using RDMA-based gRPC implementations and/or treating gRPC as just a coordination layer for synchronizing bulk video frame transfers between policy and \modelname over NCCL. This will remove the complexity and lossiness of frame encoding and decoding, shifting the burden to high-throughput datacenter networks.

\subsection{Session-based state}
\modelname uses the appearance of previously generated video frames to guide the appearance of subsequent ones, enabling high-quality generation of richly dynamic worlds. Internally, this is equivalent to the \modelname KV cache state. At the integration level, we abstract it as sessions --- each rollout starts with a new session request, which includes the (seed) first frame of simulation and the map representation of the world.

The \modelname server generates a fresh ID for the session and associates it with the received inputs and a pre-allocated KV cache. It returns the ID to the AlpaSim client, which then includes it in all its requests, indicating which rollout state each request corresponds to. This is an established pattern already in use for other stateful services, such as policy and traffic simulation.

\subsection{Supporting chunked generation}
Each frame of video generated in a single autoregressive step is conditioned on a world-scenario render (see \cref{sec::conditioning} and \cref{fig:hdmap_example}). As the generation is chunked, the simulator needs to provide the ego and actor poses for all frames in the chunk when making the request. This is challenging because the policy and traffic models cannot mutate these trajectories mid-way through a video chunk without invalidating the entire video chunk. While the goal for future generations of world models is to reduce the chunk size and eventually achieve frame-at-a-time generation, we have considered two strategies for handling the current chunk-based semantics and have implemented one in AlpaSim.

\subsubsection{Post-fetch generation}
One strategy, termed post-fetch generation, is to roll out the policy and traffic models throughout a video model chunk without providing the policy with new visual inputs. This requires the policy to support generation when the last available frame lags behind the last available ego position, a property common in production AV stacks but less so in research models. Once the policy and traffic calls within the chunk have been rolled out, a video-model request is made using the freshly generated trajectories. The response video frames logically occur \emph{before} the video model request; hence, injecting them into the simulation timeline puts the timeline out of order. Because video chunk fetch occurs after advancing the simulation timer, we call this strategy ``post-fetch''.

\subsubsection{Pre-fetch generation}
In AlpaSim, instead of requiring out-of-sync video \& egomotion processing, we choose to have a single pass of the policy and traffic models generate multi-step \emph{trajectories}, rather than just single-step next-pose prediction. This also applies to physics and controller modules that operate at the trajectory level rather than the pose level.

In the simulation loop, we enforce that both services make their predictions at the same timestamps as the video model chunk boundaries, with higher priority. Each time a new chunk starts, we append the newly generated trajectories to the ego's and the actor's historical trajectories, committing to follow them until the next chunk starts. We then interpolate those trajectories at instants corresponding to video frames and send a request to the video model. We convert each frame in response to an event that injects it into the rollout state at the precise timestamp when it would logically be captured. We call this strategy pre-fetch because the video model call happens before advancing the simulation timer. The choice of pre-fetch for AlpaSim is mainly to preserve the ordering of events, which could lead to elusive bugs if violated.

\section{Post-training \modelname as a World-Action Model (WAM)}
\label{sec::policy_finetune}

A complementary question 
is whether the representations \modelname learns during generative pre-training are themselves useful for \emph{driving}: can the same backbone be repurposed into a competitive policy?
We fine-tune the single-view checkpoint (\modelnamesv, 2\,B parameters, 120$^\circ$ front-wide camera) into an end-to-end trajectory predictor and compare it against Alpamayo~1.5~\citep{nvidia2026alpamayo} inside the same AlpaSim closed-loop stack.
Because this task does not require world-scenario map conditioning control, we initialize our policy prediction training using an \modelname causal autoregressive checkpoint prior to world-scenario map finetuning.
The policy predicts a $6.4$-second future ego trajectory at $10$\,Hz ($64$ waypoints).

\subsection{Architecture and training}
The \modelname causal DiT backbone is kept structurally identical, with two additions per frame.
First, patch features from a DINOv2 (\texttt{dinov2\_vitb14})~\citep{oquab2023dinov2}-initialized encoder and from a $30^\circ$ front-telescope camera are passed through learnable linear projections and summed into the noisy video latent tokens as additional conditioning.
Second, a \emph{history} token (embedding the previous $1.6$\,s of ego motion via a small MLP) is appended to each frame's token set and interleaved with the video patches along the temporal axis.
The attention mask preserves the original video-to-video causality (a video token attends only to current and past video tokens, exactly as during generative pre-training) and extends it to history tokens with a one-way coupling: each history token attends to all current and past video tokens but not to any other history token.
Let $h$ denote the DiT output at a history-token position. This $h$ is fed to a $12$-layer U-Net-shaped MLP that parameterizes a flow matching velocity field $\mathbf{u}{\phi}(\boldsymbol{\tau}_s, s;\,h)$ over the future-trajectory latent $\boldsymbol{\tau}\in\mathbb{R}^{64\times 3}$, using the same flow matching formulation as the video backbone (\cref{sec::training}).

We train with a joint denoising objective summing flow matching losses on the video and trajectory latents, with \emph{independently sampled} flow-matching times (per-frame $t$ for video, per-sequence $s$ for trajectory) so that the history token must extract information at whatever fidelity of visual context is currently available, rather than co-adapting to a single noise level.
At inference we drop the interleaved-noise schedule and run a single causal forward pass: $4$ video latent frames with small noise, one history token, and $4$ flow-matching steps to sample the trajectory. Since we only care about predicting the future trajectory, no video denoising is needed, so the DiT backbone runs only once and the $4$ denoising steps are confined to the lightweight trajectory MLP.

\subsection{Closed-loop comparison to Alpamayo~1.5}
We evaluate the same \modelname WAM checkpoint under the original Alpamayo~1.5 protocol: a $10$\,Hz replanning rate and 20-second rollouts over $574$-scene \emph{Physical~AI Autonomous Vehicles NuRec} dataset~\citep{nvidia2025physicalaiav}. 
The subsetting here is only done to exclude scenes on which \modelname WAM was trained on.
Compared to Alpamayo~1.5, the \modelname-finetuned policy reduces \texttt{Collision} from $6.9\%$ to $4.2\%$ (\texttt{Front}: $1.0\%$ to $0.9\%$, \texttt{Lateral}: $0.6\%$ to $0.4\%$, and \texttt{Rear}:$5.3\%$ to $3.0\%$), despite using roughly five times fewer parameters ($\sim$2\,B vs.\ $\sim$10\,B).
The same backbone that synthesizes next-frame camera observations therefore encodes scene state at a fidelity sufficient to outperform a dedicated, much larger policy.
World-model--based policies have previously been explored in domains such as robotic manipulation~\citep{ye2026world}, and here we show that the same paradigm has the potential to transfer \emph{to driving}---a setting where the future is highly multimodal, the safety bar is high, and policies must generalize across a long tail of real-world traffic situations.
For autonomous driving in particular, this suggests that the \modelname backbone need not remain a passive sensor simulator wrapped around an external driving stack: the same model can act as the shared backbone of the driving policy itself, and we see joint world-model--policy training for AV, where one model learns to both render and drive through real-world traffic, as a promising direction for future work. \cref{sec::closed_loop_eval} and \cref{fig:closed_loop_sim} show additional results on cross-policy comparison in both Nurec and \modelname simulators. 

\begin{figure}[t!]
    \centering
    \includegraphics[width=0.98\linewidth]{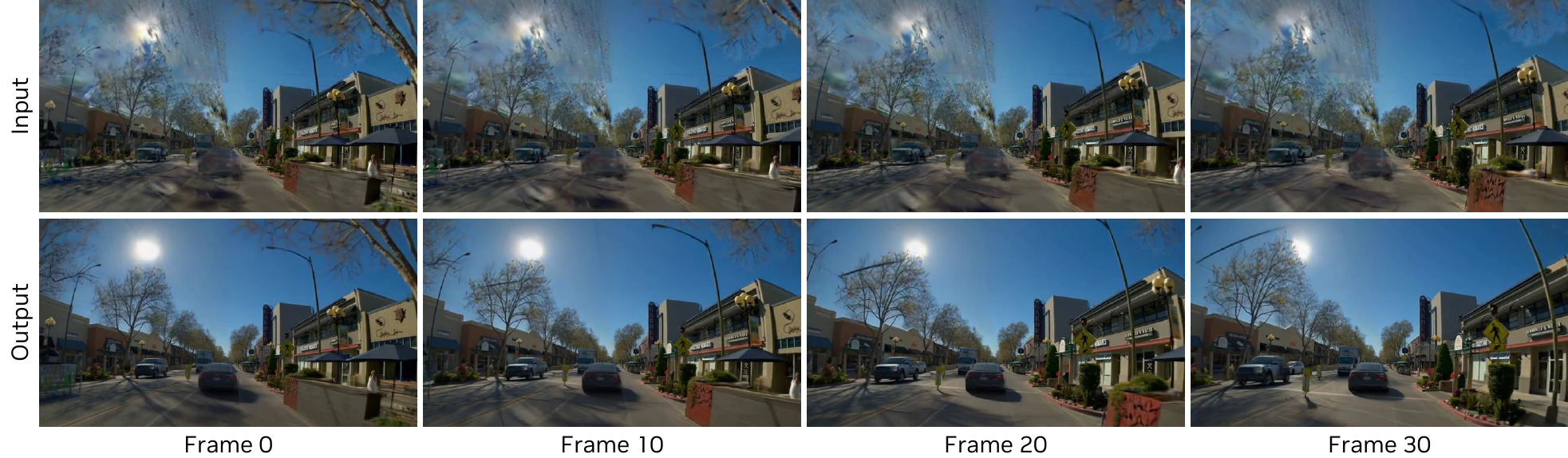}
    \caption{Reconstruction artifact correction. The first row shows input frames rendered by a reconstruction-based simulator, which can contain novel-view artifacts such as blur, ghosting, missing regions, and spurious geometry. The second row shows the corresponding OmniDreams-corrected outputs. The post-trained OmniDreams model removes reconstruction artifacts while preserving the scene layout, camera viewpoint, and driving-relevant structures.}
    \label{fig:fixer}
\end{figure}

\section{Post-Training OmniDreams as a Diffusion Fixer}

Beyond serving as a standalone generative sensor simulator, OmniDreams can also be used to improve existing reconstruction-based simulation systems. Neural reconstruction methods, including NeRFs and 3D Gaussian Splatting, provide an effective way to build simulation environments directly from real-world sensor data and have become a practical foundation for autonomous-driving simulation. However, these systems remain sensitive to novel-view extrapolation. When the requested camera pose deviates substantially from the original capture trajectory, reconstructed renderings can exhibit blurred details, missing content, ghosting, floating geometry, and other view-dependent artifacts. These artifacts can degrade the realism of closed-loop rollouts and may introduce undesirable distribution shifts for downstream policies.

To address this limitation, we post-train a distilled autoregressive OmniDreams checkpoint for reconstruction-artifact correction. We construct paired training data consisting of degraded rendering from a reconstruction-based simulator and corresponding clean target images. Inspired by prior works ~\citep{wu2025difix3d+,de2026artifixer, zhang2026diffusionharmonizer}, during post-training, rather than starting the denoising process from random Gaussian noise, we start the denoising process from the degraded rendering itself, following the formulation used in prior works. In this setting, OmniDreams learns a flow that maps artifact-corrupted reconstruction outputs onto the clean-image manifold while preserving the scene layout, camera viewpoint, and driving-relevant structure.

At inference time, this post-trained model can be applied as an autoregressive correction module on top of pre-reconstructed scenes. Given a rendered frame from the reconstruction system, OmniDreams refines the image by removing novel-view artifacts while maintaining temporal consistency through its causal history and KV-cache conditioning. This enables reconstruction-based simulators to retain their strengths in scene-specific geometric grounding, while benefiting from the learned visual priors of a generative world model. We visualize qualitative results in Fig.~\ref{fig:fixer}.

\section{Experiments and Results}
\label{sec::results}

To be effective, a generative closed-loop simulator must accurately model real-world behavior and appearance while maintaining simulation capabilities that extend beyond those of standard neural reconstruction approaches. This section details how \modelname delivers on this premise across four main axes: generating high-quality appearance and realistic dynamics, producing stable rollouts over extended horizons, enabling the controlled generation of long-tail scenarios, and facilitating real-time closed-loop policy evaluation.

\subsection{Simulation Quality}

\begin{figure}[t]
    \centering
    \includegraphics[width=0.98\linewidth]{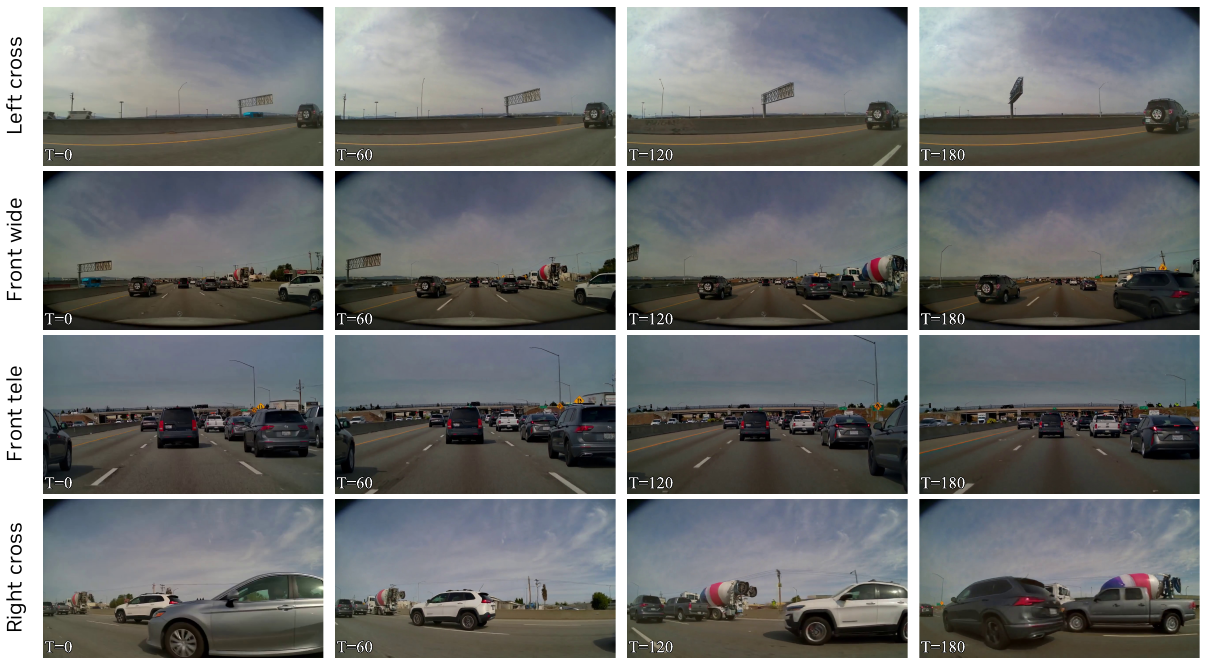}
    \caption{Multi-view generation.
    \modelname generates synchronized observations for multiple camera views from a shared scene state. Across cross-left, front-wide, front-tele, and cross-right cameras, road layout, dynamic actors, lighting, and scene context remain coherent over time, enabling closed-loop policies that consume multi-view camera inputs.}
    \label{fig:multiview_generation}
\end{figure}

\modelname generates physically realistic scene dynamics and traffic cues that directly affect AV decision-making. This includes the articulated motion of vulnerable road users (VRUs) and windshield wipers, phenomena that remain notoriously challenging for standard neural reconstruction approaches.

The model also preserves critical semantics and control cues within the scene, including traffic lights, lane markings, traffic signs, and precise roadway structures. This baseline fidelity is vital, as structural rendering errors in these details inevitably propagate into downstream policy behavior.

Quantitative metrics in this subsection are evaluated on 1{,}000 clips sampled from the 5{,}000-clip held-out evaluation split of the RDS-HQ-1M dataset.

In \cref{tab::training_stages} we compare different stages of the \modelname training for \modelnamesv. The final distilled \modelname model achieves the best FVD score (StyleGAN-V \citep{stylegan-v}) and the strongest fidelity to conditioning signals, as measured by 3D detection metrics \citep{li2022bevformerlearningbirdseyeviewrepresentation, hung2024} and lane-line regression and detection scores \citep{luo2023latr3dlanedetection}, as well as the Temporal Sampson \citep{sampson} score, which measures cross-frame consistency. The final model remains close to the bidirectional model while enabling real-time causal generation.
This is partially due to the high-quality distillation dataset we curated.
In \cref{tab::lightvae}, we show that swapping to the LightVAE decoder (discussed in \cref{sec::inf_model} to optimize for fastest inference) comes with a corresponding quality trade-off visible in simulation quality metrics. 

\begin{table}[h]
  \centering
  \small
  \setlength{\tabcolsep}{4pt}
    \resizebox{\linewidth}{!}{%
    \begin{tabular}{lcc ccc ccc}
      \toprule
        & \multicolumn{2}{c}{Generation quality}
        & \multicolumn{3}{c}{3D vehicle detection (BEVFormer)}
        & \multicolumn{3}{c}{Lane line (LATR)} \\
      \cmidrule(lr){2-3} \cmidrule(lr){4-6} \cmidrule(lr){7-9}
      Training stage
        & FVD $\downarrow$ & Temp.\ Sampson $\downarrow$
        & LET-AP $\uparrow$ & LET-APL $\uparrow$ & LET-APH $\uparrow$
        & F1 $\uparrow$ & $x$-err.\ (far) $\downarrow$ & Cat.\ Acc.\ $\uparrow$ \\
      \midrule
      Bidirectional (AV adapted)          & 26.8          & \textbf{1.83} & 0.378          & 0.240          & 0.366          & 0.823          & 0.337          & 0.957          \\
      Causal (Diffusion Forcing)      & 31.7          & 1.87          & 0.221          & 0.136          & 0.214          & 0.775          & 0.418          & 0.941          \\
      Distilled (Self Forcing)            & \textbf{24.8} & 1.90          & \textbf{0.400} & \textbf{0.255} & \textbf{0.388} & \textbf{0.828} & \textbf{0.313} & \textbf{0.961} \\
      \bottomrule
    \end{tabular}%
    }
  \caption{Training-stage comparison for \modelname across the bidirectional abstract state-conditioned model, the causal many-step Diffusion Forcing student, and causal few-step Self Forcing students. \emph{Generation quality} is measured directly on the synthesized frames (FVD against real; Temporal Sampson for geometric consistency across frames). \emph{3D detection} metrics LET-AP/APL/APH \cite{hung2024} are computed by running the off-the-shelf BEVFormer detector~\citep{li2022bevformerlearningbirdseyeviewrepresentation} on \modelname outputs. Similarly, \emph{lane line} F1, x-coordinate rMSE and accuracy are computed by running the off-the-shelf LATR lane detector~\citep{luo2023latr3dlanedetection} on \modelname outputs.}
  \label{tab::training_stages}
\end{table}

\begin{table}[h]
  \centering
  \small
  \setlength{\tabcolsep}{4pt}
    \resizebox{\linewidth}{!}{%
    \begin{tabular}{lcc ccc ccc}
      \toprule
        & \multicolumn{2}{c}{Generation quality}
        & \multicolumn{3}{c}{3D detection (BEVFormer)}
        & \multicolumn{3}{c}{Lane line (LATR)} \\
      \cmidrule(lr){2-3} \cmidrule(lr){4-6} \cmidrule(lr){7-9}
      Training stage
        & FVD $\downarrow$ & Temp.\ Sampson $\downarrow$
        & LET-AP $\uparrow$ & LET-APL $\uparrow$ & LET-APH $\uparrow$
        & F1 $\uparrow$ & $x$-err.\ (far) $\downarrow$ & Cat.\ Acc.\ $\uparrow$ \\
      \midrule
      Distilled (Original VAE)            & \textbf{24.8} & \textbf{1.90}          & \textbf{0.400} & \textbf{0.255} & \textbf{0.388} & \textbf{0.828} & \textbf{0.313} & \textbf{0.961} \\
      Distilled (LightTAE decoder)        & 45.4          & 2.02          & 0.376          & 0.237          & 0.365          & 0.813          & 0.352          & 0.952          \\
      \bottomrule
    \end{tabular}%
    }
  \caption{We show a tradeoff between decoder latency and generation quality.}
  \label{tab::lightvae}
\end{table}

\Cref{fig:multiview_generation} shows a representative four-camera rollout with synchronized cross-left, front-wide, front-telescope, and cross-right views from our \modelnamemv. Depending on the desired configuration, users can adapt the model by post-training for the relevant camera count and type. While each camera increases inference compute, real-time performance is observable on up to four cameras on NVIDIA GB300.

\subsection{Long Rollouts}

\begin{figure}[ht]
    \centering
    \includegraphics[width=0.98\linewidth]{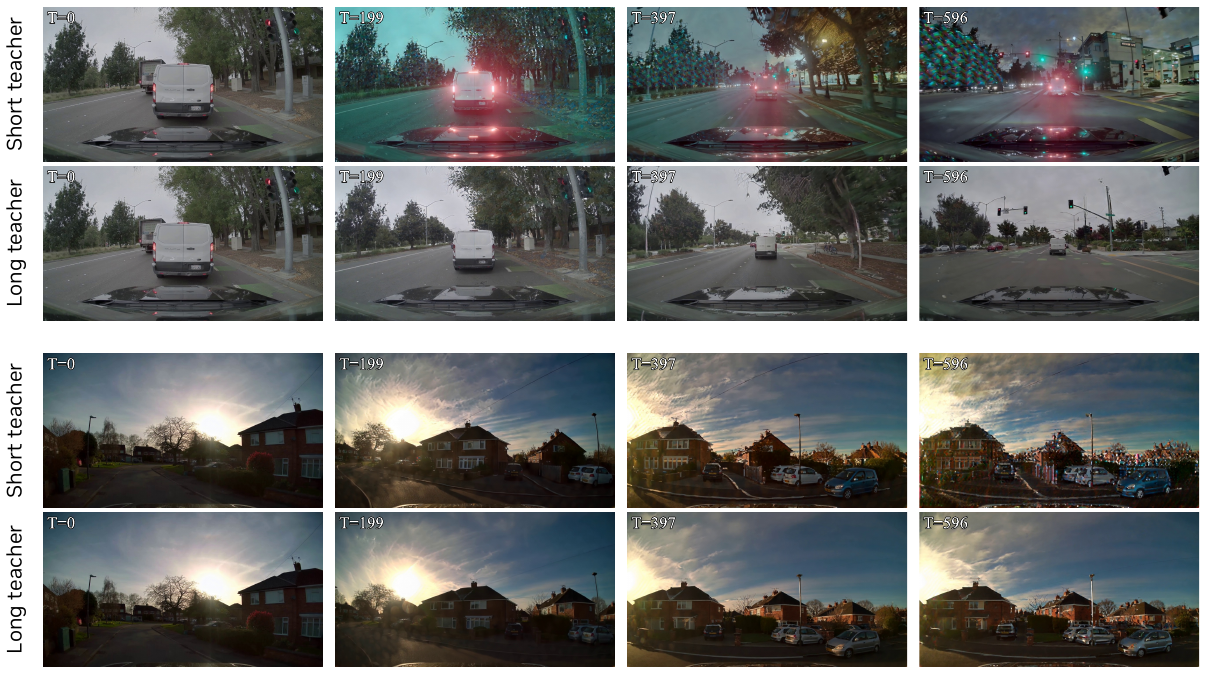}
    \caption{Progressive teacher strategy for long rollouts.
    Within each two-row group, the top row shows the model after Self Forcing distillation with a short-context teacher, while the bottom row shows the same model after continued distillation with a long-context bidirectional teacher (\cref{sec::training}). Frames are sampled at $T = \{0,\,199,\,397,\,596\}$ along $597$-frame autoregressive rollouts ($\sim$20\,s at 30\,FPS). When the rolling KV cache extends beyond the short teacher's training context, the top-row rollouts accumulate shifting artifacts; the progressive long-teacher stage substantially reduces these artifacts and better preserves scene structure and object identity over time.}
    \label{fig:long_rollout}
\end{figure}

Stable long-horizon generation is enabled by three design choices introduced earlier: a causal autoregressive formulation with a streaming KV cache (\cref{sec::model}), Self Forcing distillation with a progressive long-context teacher to suppress exposure bias and shifting artifacts (\cref{sec::training}), and a bounded local-attention window with attention-sink tokens (tokens from the first RGB frame) at inference time. Together, these let \modelname recondition to an updated state and recent history at every step, while avoiding appearance drift, identity loss, and the accumulation of artifacts that typically affect autoregressive video models over long rollouts. \Cref{fig:long_rollout} shows representative 20-second autoregressive rollouts and the clear improvement in visual quality and temporal stability obtained by continuing Self Forcing distillation with the progressive long-context teacher. In practice, we find the model can maintain decent quality even after minutes of rollouts.

We found using long-context teacher to continue Self Forcing (discussed in \cref{sec::distillation}) is key to the success of high-quality long rollout, shown in \cref{tab:segmented_fvd_long_rollouts}. We quantify this long-horizon behavior with segmented FVD. Each 20-second rollout is split into four five-second windows, and each window is compared against the same real-video front-wide reference distribution. 

\begin{table}[h]
  \centering
  \small
  \setlength{\tabcolsep}{6pt}
  \begin{tabular}{l c c c c c c}
    \toprule
    Training teacher
      & 0--5\,s $\downarrow$
      & 5--10\,s $\downarrow$
      & 10--15\,s $\downarrow$
      & 15--20\,s $\downarrow$
      & Mean $\downarrow$
      & $\Delta$ $\downarrow$ \\
    \midrule
    Short-context teacher
      & 109.3 & 183.0 & 258.3 & 409.2 & 240.0 & 299.9 \\
    Progressive long-context teacher
      & \textbf{95.5} & \textbf{151.0} & \textbf{202.5} & \textbf{268.4} & \textbf{179.4} & \textbf{172.9} \\
    \bottomrule
  \end{tabular}
  \caption{Segmented FVD on 20-second rollouts, each rollout is split into four temporal windows and compared against the same precomputed real-video front-wide reference distribution; lower is better. $\Delta$ is the final-window FVD minus the first-window FVD.}
  \label{tab:segmented_fvd_long_rollouts}
\end{table}

\subsection{Long-tail Coverage and Counterfactual Variations}

\modelname provides two complementary handles for accessing scenarios that are rare or impossible to capture in fleet logs. First, the prompt and first frame-driven conditioning enables a single trained checkpoint to generate broad variations in environments, agents, and ego behavior without retraining (\cref{sec::editing}). Second, lightweight targeted finetuning steers the model toward distributions that are otherwise too rare to emerge from natural data sampling (\cref{sec:ood_obj}). At scale, both handles turn the simulator into a data engine for balanced training and evaluation sets with broad scenario coverage.

\subsubsection{Controllable Scenario Editing}
\label{sec::editing}

The conditioning interfaces introduced in \cref{sec::model} expose two channels for counterfactual editing: the \emph{text prompt}, which controls environmental attributes such as weather, lighting, and time of day~\citep{wu2025chronoedit}; and the \emph{abstract world-scenario map} (lane geometry and bounding boxes), which controls scene structure and agent layout. Together with the first-frame RGB seed, these channels allow a developer to systematically vary what the policy sees while keeping unspecified attributes stable across variants.

\begin{figure}[ht]
\centering
\includegraphics[width=0.98\linewidth]{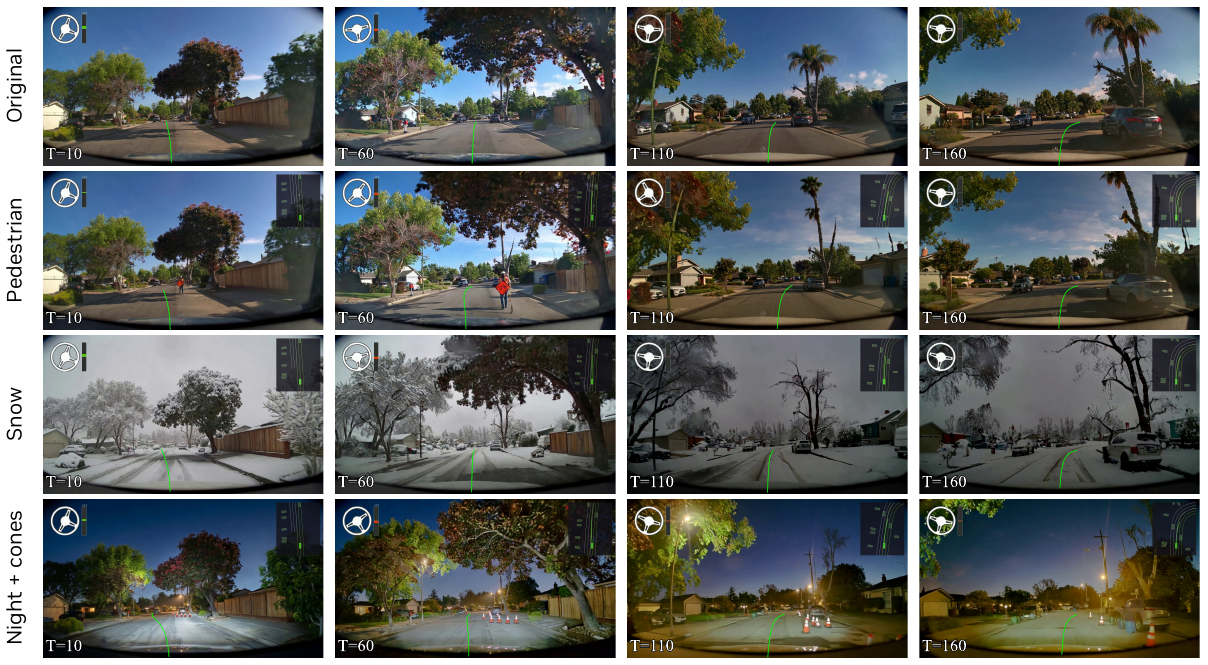}
\caption{Controllable scenario editing.
Starting from a given scene, \modelname supports targeted edits by modifying its conditioning inputs: text prompts can be used to modify appearance attributes such as weather and time of day of the first frame condition, and by also adapting the abstract world-scenario map, and ego-state conditioning changes the driven trajectory. Unedited attributes such as road geometry and distant scene context remain visually stable across variants.}
\label{fig:editing}
\end{figure}

\Cref{fig:editing} illustrates these controls on a single seed scene. By editing the conditioning signals such as the first frame, text prompt, and abstract world-scenario map, we enable targeted scenario variations. These edits allow users to evaluate the policy's behavior in different scenarios. Across these variants, attributes that are not targeted by the edit, including road geometry, static scene structure, and distant background content, remain visually stable.

\subsubsection{Out-of-Distribution Object Modeling}
\label{sec:ood_obj}

Beyond modifying weather, lighting, ego motion, and in-distribution traffic participants, \modelname can also be adapted to model objects that are out-of-distribution relative to standard autonomous-driving logs. This capability is important for stress-testing policies under unusual visual conditions, such as the sudden appearance of uncommon animals, oversized objects, or other rare obstacles that are difficult to collect in real-world fleet data at scale.

A naive approach is to directly edit the first-frame RGB image by inserting out-of-distribution objects and then generate the full video. However, this often leads to visual artifacts and inconsistent dynamics, as the inserted objects are not represented in the world-scenario map and therefore conflict with the dynamic cuboid in the world-scenario map  conditioning used by the model. As a result, the model receives inconsistent signals: the RGB image suggests the presence of an object, while the structured conditioning provides no corresponding position, extent, or trajectory. This mismatch makes it difficult for the model to reconcile the edited appearance with the simulator state over time.

\begin{figure}[ht]
    \centering
    \includegraphics[width=0.98\linewidth]{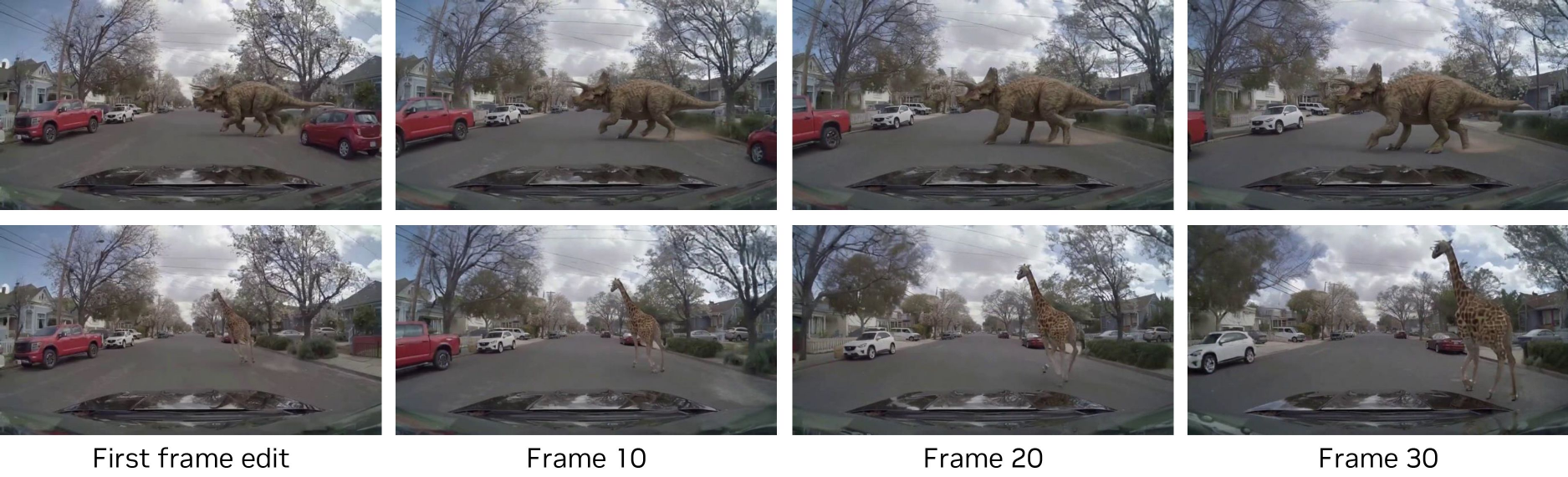}
    \caption{Out-of-distribution object modeling. Given a first-frame edit that inserts an uncommon object into the scene, \modelname generates plausible motion for the inserted object over time without an explicit cuboid trajectory, while maintaining consistency with the surrounding road geometry and scene context.}
    \label{fig:ood}
\end{figure}

To improve flexibility, we further post-train another variant of \modelname with randomized dynamic-cuboid dropout. During this stage, cuboids representing dynamic agents are randomly removed from the abstract world-scenario map, while static cuboids, such as parked vehicles, are left unchanged. This prevents the model from relying exclusively on dynamic cuboids to determine the location and motion of every moving object. Instead, the model learns to infer plausible object persistence and motion from visual history, the first-frame seed, and surrounding scene context. At inference time, this allows out-of-distribution objects inserted into the initial frame to propagate more naturally through the generated video, even when they lack an explicit cuboid trajectory in the conditioning signal. We visualize the results in \cref{fig:ood}.

\subsection{Closed-Loop Evaluation}
\label{sec::closed_loop_eval}

\begin{figure}[ht]
    \centering
    \begin{subfigure}[b]{0.30\linewidth}
        \centering
        \includegraphics[width=\linewidth]{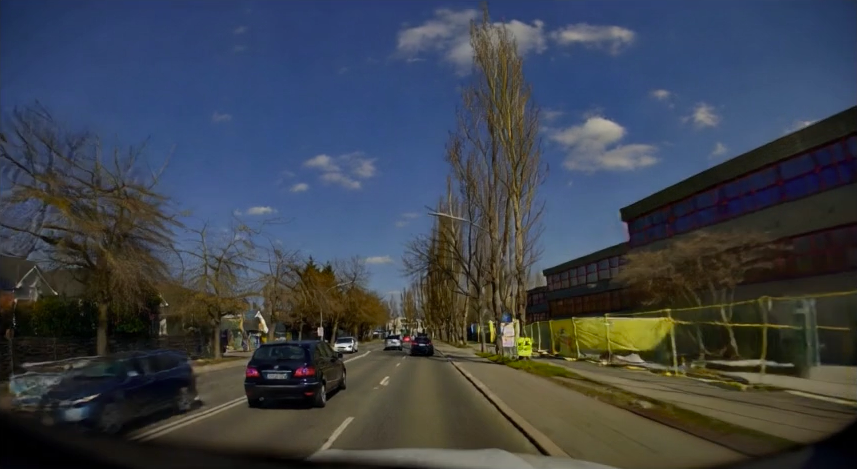}
    \end{subfigure}
    \hfill
    \begin{subfigure}[b]{0.165\linewidth}
        \centering
        \includegraphics[width=\linewidth]{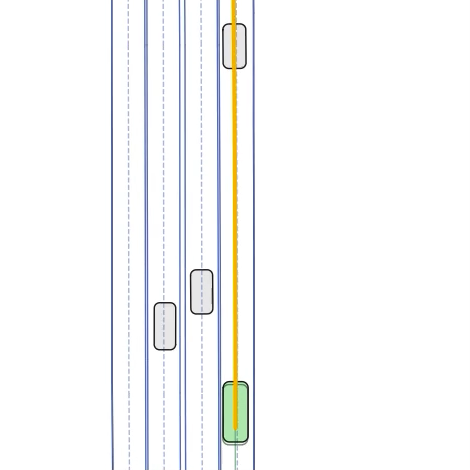}
    \end{subfigure}
    \hfill
    \begin{subfigure}[b]{0.30\linewidth}
        \centering
        \includegraphics[width=\linewidth]{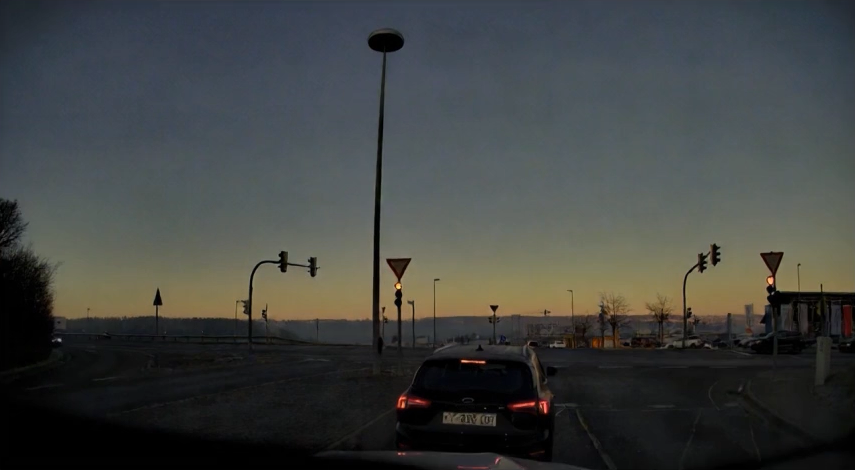}
    \end{subfigure}
    \hfill
    \begin{subfigure}[b]{0.165\linewidth}
        \centering
        \includegraphics[width=\linewidth]{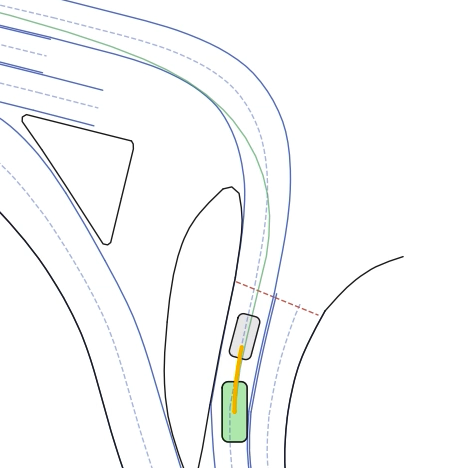}
    \end{subfigure}
    \caption{Screenshots of closed-loop simulation driven by \modelname. Each (camera, BEV) pair shows one rollout: the camera observation synthesized by \modelname, paired with AlpaSim's bird's-eye-view debug overlay. The BEV visualizes the abstract world-scenario map condition and tracked objects, with the ego vehicle in green and its predicted trajectory from the Alpamayo 1 policy drawn as a yellow line.}
    \label{fig:closed_loop_sim}
\end{figure}

A central application of \modelname is to act as a generative sensor simulator inside a closed-loop AV evaluation stack, as shown in \cref{fig:closed_loop_sim}.
We pair it with NVIDIA AlpaSim~\citep{nvlabs2025alpasim} as the simulation orchestrator (\cref{sec::applications}) and Alpamayo~1.5~\citep{nvidia2026alpamayo} as the driving policy, so that each closed-loop step alternates between policy action and \modelname sensor synthesis. 
We ask two concrete questions in this setup. 
First (\cref{sec::closed_loop_compare}), how does the generative simulator compare against a reconstruction-based one when both drive the same policy through the same closed-loop stack? Second (\cref{sec::policy_finetune}), are the representations \modelname has learned during generative pre-training themselves useful for driving---i.e.\ can the same backbone be fine-tuned into a competitive policy?

\subsubsection{Closed-Loop Comparison to Reconstruction-based NuRec Simulator}
\label{sec::closed_loop_compare}

We swap only the sensor simulator between NVIDIA NuRec~\citep{nvidia2024nurec}, the reconstruction-based simulator that AlpaSim ships with by default, and \modelname; the details of the AlpaSim integration are described in \cref{sec::applications}. The orchestrator (AlpaSim), the traffic and physics services, and the per-scene initial state are held fixed; differences in the resulting rollout metrics therefore isolate the sensor simulator's influence on policy behavior. To check that this conclusion is not specific to a single policy, we sweep across multiple policy classes rather than fixing one.

\paragraph{Policies compared.}
We evaluate four policy classes inside the same AlpaSim closed-loop stack. The first is the \modelname-derived World-Action Model (\emph{\modelname WAM}); we describe its architecture and training in detail in \cref{sec::policy_finetune} and treat it here purely as another policy under test. The remaining three are camera-input variants of Alpamayo~1.5~\citep{nvidia2026alpamayo}: the full \emph{Alpamayo~1.5} configuration uses all four trained cameras (front-wide, front-telescope, cross-left, cross-right); since Alpamayo~1.5 is trained to accept any subset of these views, we additionally treat \emph{Alpamayo~1.5 (2 cam)} (front-wide + front-telescope only) and \emph{Alpamayo~1.5 (1 cam)} (front-wide only) as distinct policy classes by restricting the cameras provided at inference time.

\paragraph{Evaluation set and metrics.}
NuRec consumes per-scene 3D Gaussian Splatting reconstructions, while \modelname consumes a first-frame RGB seed plus a world-scenario rendering (abstract state world-scenario map; see \cref{sec::conditioning}), so a direct comparison is only possible on scenes for which both representations are available. We use a 501-scene subset of the released \emph{Physical~AI Autonomous Vehicles NuRec} dataset~\citep{nvidia2025physicalaiav}.
This subset is restricted to scenes that include OmniDreams-compatible world-scenario map alongside the published NuRec reconstructions, excluding any scenes previously used in \emph{\modelname WAM}'s training.
We roll out each scene for 20 seconds in a closed loop. The policies replan every $533$\,ms (we throttle the replanning rate to \modelname's chunk rate of 533 ms).
Following the AlpaSim protocol of Alpamayo-1.5~\citep{nvidia2026alpamayo}, an incident such as a collision or off-road event is only counted when the ego is within $4$\,m of the corresponding ground-truth trajectory. Because rollouts stay close to the recorded trajectory in this regime, NuRec, which is fit to the original capture, effectively serves as a strong reconstruction-based reference of the real recording, and its closed-loop metrics give a strong reference for how each policy would behave in the real world.
We report per-policy lower-is-better metrics: the headline summary \emph{All Incidents (Collision + Offroad)}, \emph{Collision (Front)}, \emph{Collision (Lateral)}, \emph{Collision (Rear)}, and \emph{Offroad}.

\begin{figure}[t]
    \centering
    \includegraphics[width=0.98\linewidth]{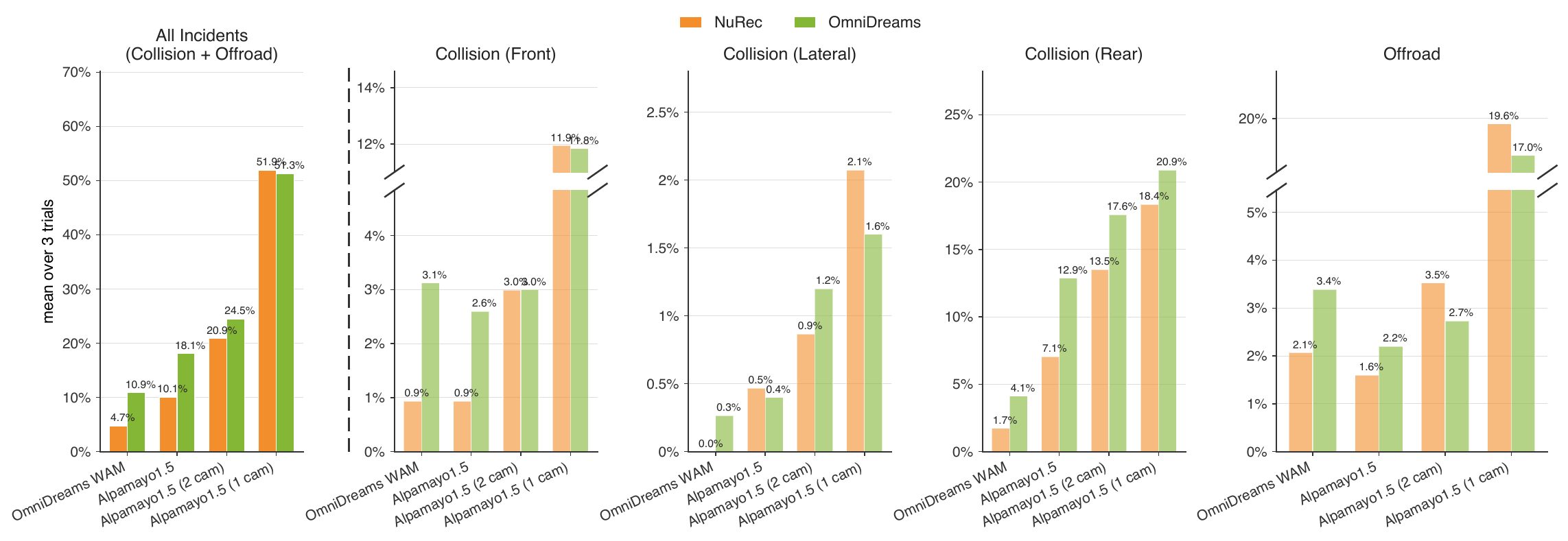}
    \caption{Closed-loop comparison between NuRec and \modelname across multiple policy classes.
    Each pair of bars compares the same policy when AlpaSim's sensor simulator is NuRec (orange) versus \modelname (green), averaged on the 501-scene subset of \emph{Physical~AI Autonomous Vehicles NuRec}; lower is better. From left to right within each panel: \modelname WAM (see \cref{sec::policy_finetune}), Alpamayo~1.5 (4 cameras), Alpamayo~1.5 (2 cam, front-wide + front-telescope), and Alpamayo~1.5 (1 cam, front-wide only). The headline \emph{All Incidents} panel (left) preserves the policy ranking when switching from NuRec to \modelname, indicating that \modelname is a faithful proxy for closed-loop policy evaluation. The remaining panels break this summary down by incident type.}
    \label{fig:closed_loop_renderer}
\end{figure}

\paragraph{Results and implications.}
\Cref{fig:closed_loop_renderer} summarizes closed-loop incident metrics.
The headline \emph{All Incidents} panel (leftmost) shows that the relative ranking of policies under NuRec is preserved under \modelname: from \modelname WAM as the strongest policy through Alpamayo~1.5 (4 cam), Alpamayo~1.5 (2 cam), and finally Alpamayo~1.5 (1 cam) as the weakest. Combined with the observation above that NuRec is a strong reconstruction-based proxy for the real recording in this regime, the preservation of this ranking under \modelname is direct evidence that \modelname is itself a faithful proxy of the real world for closed-loop policy evaluation -- a developer comparing policy candidates inside \modelname would draw similar conclusions as if comparing them against logged data.

\paragraph{Visual Realism}

\begin{figure}[ht]
    \centering
    \includegraphics[width=0.54\linewidth]{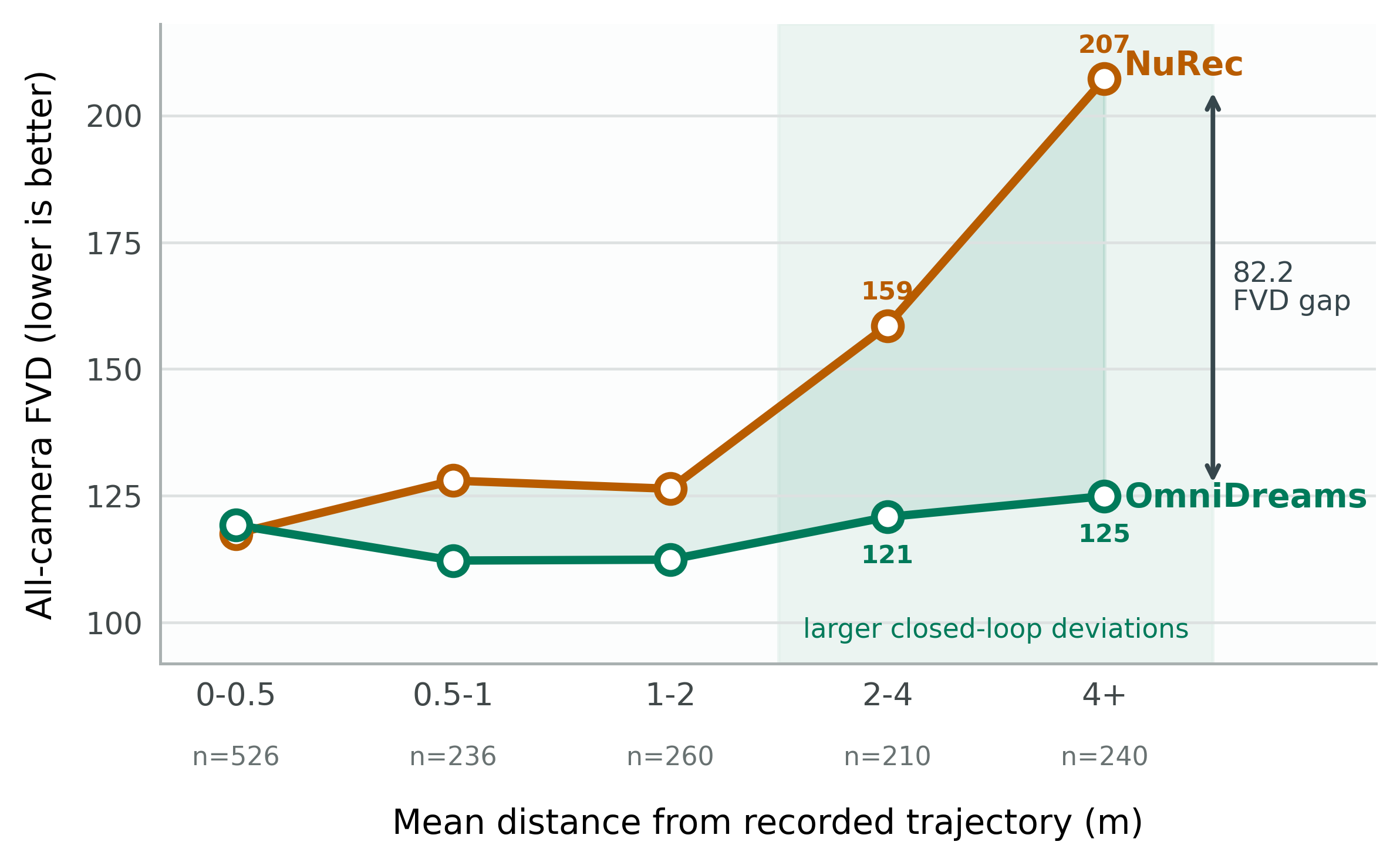}
    \caption{Video realism under increasing closed-loop trajectory deviation.
    We generate ego trajectories by running policies in AlpaSim closed loop, freeze those trajectories, and replay each one through both NuRec and \modelname. We then bin the resulting four-camera clips by their mean distance from the recorded ego trajectory and compute FVD against the corresponding recorded-video distribution; lower is better. Near the recorded trajectory the two simulators are comparable, but as rollouts move farther from the capture path, NuRec degrades sharply while \modelname remains substantially more stable.}
    \label{fig:omnidreams_vs_nurec}
\end{figure}

Incident rates show that \modelname preserves policy rankings, but closed-loop evaluation also depends on whether the simulator can keep producing realistic sensor observations after the policy deviates from the recorded trajectory. Closed-loop simulation is useful precisely because the ego trajectory is no longer fixed to the log: the policy selects actions, those actions move the vehicle, and the simulator must synthesize the sensor observations induced by the resulting trajectory. To evaluate this regime directly, we take closed-loop trajectories generated under both the NuRec and \modelname sensor backends, replay each trajectory through both simulators, and group the resulting videos by their mean distance from the recorded ego trajectory. \Cref{fig:omnidreams_vs_nurec} shows that NuRec and \modelname are comparable near the recorded trajectory, where a reconstruction-based renderer remains close to its capture path. As the mean trajectory deviation increases, however, NuRec's FVD rises rapidly while \modelname remains much more stable.

This is precisely where reconstruction-based renderers are stressed: novel viewpoints, changes in optical flow, and shifted occlusion boundaries require extrapolating beyond the captured rays. In contrast, \modelname can use its learned video prior and world-scenario conditioning to synthesize plausible observations beyond the recorded camera path.

\begin{figure}[ht]
    \centering
    \includegraphics[width=0.98\linewidth]{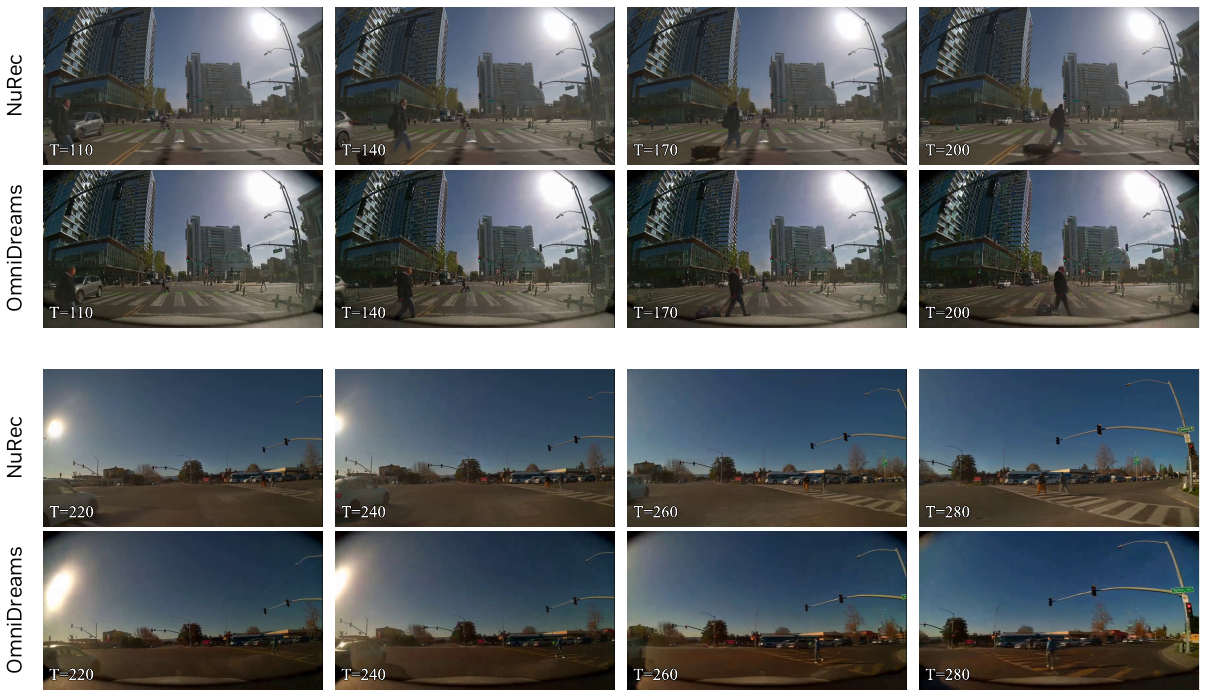}
\caption{Qualitative comparison to NuRec.
NuRec renders the scene from a per-scene reconstruction, while \modelname generates the camera observation from a first-frame seed and abstract world scenario. \modelname preserves key driving cues such as lane geometry, traffic participants, and scene layout while improving pedestrian quality and producing more natural pedestrian motion that is difficult for reconstruction-based simulators to capture.}
\label{fig:omnidreams_nurec_comp}
\end{figure}

OmniDreams is also better suited for rendering dynamic objects. \Cref{fig:omnidreams_nurec_comp} further illustrates the qualitative comparison between \modelname and reconstruction-based NuRec on challenging dynamic content.

\paragraph{Controllability and Compute}

NuRec requires synchronized multi-camera capture and an offline 3DGS reconstruction pipeline per scene, and inherits known artifacts in regions of low view overlap or off-capture-path viewpoints. At the same time, \modelname only needs a calibrated first-frame seed and the abstract world-scenario representation which can be produced by the AV data pipeline (\cref{sec::data_sources}, \cref{sec::conditioning}). One advantage is that the world-scenario map exposes the controllability and incident-coverage handles described in \cref{sec::editing} that a reconstruction-based simulator cannot natively provide.

However, a video-generation based simulator like OmniDreams natively requires much more compute than a reconstruction-based simulator, exposing a tradeoff between quality and compute.

\section{Related Work}
\label{sec::related_work}

Apart from the policy model itself, closed-loop autonomous-vehicle simulation primarily relies on two key components: a world model that synthesizes sensor observations at a requested position and time, and a simulation infrastructure that orchestrates the simulation loop and connects the policy to the world model. We organize the related work around these two axes, focusing on the topics most relevant to our approach. First, we discuss world models, including reconstruction-based neural simulators, general video generative models, and generative driving models. Second, we review streaming and orchestration infrastructure for video-diffusion-based simulation. Together, these areas provide the technical context for the design choices underlying \modelname.

\subsection{World models for closed-loop AV Simulation}

\subsubsection{Reconstruction-based World Models}
Reconstructing simulation environments from captured sensor data provides a fast and scalable alternative to traditional artist-generated environments. By recovering scenes directly from real-world observations, this approach reduces the simulation-to-reality domain gap, scales with the amount of captured data rather than the number of artists, and enables rapid turnaround from on-road events to simulation-ready environments. NeRF-\citep{mildenhall2020nerf, mueller2022instant} and 3D Gaussian Splatting~\citep{kerbl3Dgaussians, loccoz20243dgrt, wu20253dgut} based methods have become dominant approaches for high-fidelity neural reconstruction of captured driving scenarios. NeuRAD~\citep{tonderski2024neurad} is one of the first works that explicitly modeled camera and lidar sensor characteristics (rolling shutter, beam divergence, ray drop) across five automotive datasets, and EmerNeRF~\citep{yang2024emernerf} introduces self-supervised decomposition of dynamic scenes into static, dynamic, and flow fields. SiMUli~\cite{turki2026simuli} extends 3DGUT to support LiDAR rendering within 3D Gaussian Splatting formulations and addresses cross-sensor inconsistencies through a factorized 3D Gaussian representation. NVIDIA Omniverse NuRec~\citep{nvidia2024nurec} commercializes 3DGS-based reconstruction at production scale and serves as the renderer used by AlpaSim~\citep{nvlabs2025alpasim} prior to the integration described in \cref{sec::applications}. Reconstruction-based World Models synthesize faithful sensor observations within captured corridors, making them well suited for what-if testing within logged scenes. However, they extrapolate poorly to content, viewpoints, and conditions absent from the original capture, motivating the generative direction we pursue here.

\subsubsection{Video models as world simulator}
A complementary line of work frames large, general-purpose video-generation models as world simulators trained on broad, internet-scale visual data. Sora~\citep{openai2024sora} popularized this framing for diffusion-based video models; Movie Gen~\citep{polyak2024movie} extends it to longer durations and broader controllability; Wan~\citep{wan2025wan} releases an open large-scale video generative model; and Veo 3~\citep{deepmind2025veo3} adds synchronized audio and speech. Genie~\citep{bruce2024genie} takes a different framing, learning a foundation world model from unlabeled internet videos that supports action-controllable interactive rollouts from a single image, text, or sketch prompt. The NVIDIA Cosmos world-foundation-model platform~\citep{agarwal2025cosmos,nvidia2025worldsimulationvideofoundation} consolidates this line of work for the physical-AI setting and provides the backbone we post-train on. \modelname specializes Cosmos to the autonomous-driving setting: it adds action conditioning on policy trajectories and world-scenario-map control, runs in a closed loop with a real policy and traffic simulator, and meets a per-chunk latency budget that the general-purpose models in this subsection do not target.

\subsubsection{Generative AV World Models}
A parallel line of work uses video generation to synthesize driving footage conditioned on scene-level signals. 
DriveGAN~\citep{Kim2021_DriveGAN} was the first generative controllable driving system trained entirely from real-world driving logs.
More recently, DriveDreamer~\citep{wang2024drivedreamer} introduced a diffusion-based generative system for driving scenes; GAIA-1~\citep{hu2023gaia} and its multi-view successor GAIA-2~\citep{russell2025gaia2} introduce structured controllability over weather, agents, ego dynamics, and surround-view camera rigs; Vista~\citep{gao2024vista} emphasizes high-resolution generalization and multi-modal trajectory control; MagicDrive~\citep{gao2024magicdrive} and Drive-WM~\citep{wang2024drivewm} target multi-view generation with explicit 3D-geometry and BEV-style controls; and GenAD~\citep{yang2024genad} trains a generalist video predictor on a $2000+$-hour open-domain corpus. Closer to our setting, the NVIDIA Cosmos world-foundation-model platform~\citep{nvidia2025worldsimulationvideofoundation} provides the backbone we post-train, and Cosmos-Drive-Dreams~\citep{ren2025cosmosdrivedreams} shows that Cosmos-derived synthetic driving data improves downstream perception and policy training. Concurrently, Waymo introduced the Waymo World Model~\citep{waymo2026worldmodel}, which adapts a Genie-style foundation world model~\citep{bruce2024genie} to autonomous driving and generates controllable, multi-sensor (camera and lidar) rollouts conditioned on driving inputs, scene layouts, and language prompts. \modelname differs from this line on three axes: (i)~it is built for closed-loop interactivity rather than offline rollout; (ii)~it integrates a streaming KV cache and few-step distillation that drive per-chunk latency into the $100+$\,FPS regime (\cref{sec::inference}); and (iii)~it ships with a published microservice simulator (AlpaSim, \cref{sec::applications}) and a published policy stack (Alpamayo 1~\citep{nvidia2026alpamayo}).

\subsubsection{Real-time and streaming video diffusion}
Pushing per-chunk latency of video diffusion into interactive territory requires progress on both the algorithm and systems sides. Self Forcing~\citep{huang2025self} and CausVid~\citep{yin2024causvid} bridge the train--test gap of autoregressive video diffusion via self-rollout and asymmetric distillation; we adopt the Self Forcing recipe in \cref{sec::training} and combine it with distribution-matching distillation (DMD)~\citep{yin2024one}. Long-rollout stability is the focus of StreamingT2V~\citep{henschel2024streamingt2v} and FIFO-Diffusion~\citep{kim2024fifo}, and Diffusion Forcing~\citep{chen2024diffusionforcing} provides the per-token noise schedule that underpins our causal mid-training. Our streaming-attention design extends the attention-sink construction from StreamingLLM~\citep{xiao2024streamingllm} to a video-diffusion setting; ring attention~\citep{liu2024ringattention} supplies the kernel pattern for our context-parallel attention; and we discuss future extensions to sparse temporal attention~\citep{yuan2025nativesparse} and lightweight super-resolution~\citep{zhuang2025flashvsr}.

\subsection{Closed-loop AV simulation infrastructure}
Simulator design for AV research has historically operated at one of three levels of perception realism:
\begin{itemize}
\item non-visual or simple abstract graphics, usually targeting massively parallel simulation for large scale RL training, such as Waymax~\citep{gulino2023waymax} and PufferDrive~\citep{pufferdrive2025github},
\item physical and visual realism obtained through artist-generated assets, exemplified by CARLA~\citep{dosovitskiy2017carla} or MetaDrive~\citep{li2021metadrive}, 
\item graphics derived from neural reconstruction methods, marrying data-driven scenario generation with high visual quality: DriveArena~\citep{yang2024drivearena}, HUGSIM~\citep{zhou2024hugsim}, WorldEngine~\citep{opendrivelab2026worldengine} and AlpaSim~\citep{nvlabs2025alpasim}.
\end{itemize}
We believe world models open a new, 4th tier of perception realism. We demonstrate it by extending AlpaSim, as it easily admits remote, distributed video-model rendering thanks to its unique microservice-based architecture. \Cref{sec::applications} describes how we extend AlpaSim to support the chunk-based stateful semantics required by a video-diffusion world model.

\section{Conclusion}

The transition from purely reconstruction-based environments to generative world models marks a pivotal advancement in validating autonomous driving systems. By seamlessly integrating \modelname with the Alpamayo 1 policy and AlpaSim orchestrator, this framework demonstrates a robust pathway for generating highly dynamic, unconstrained testing scenarios. Because the model effectively synthesizes complex, interactive elements, ranging from rare weather anomalies to unpredictable pedestrian movements, it can directly address the long-tail edge cases that standard simulators struggle to recreate. Ultimately, this generative, closed-loop approach equips developers with the scalable tools necessary to rigorously evaluate and potentially train advanced vision-language-action policies, effectively bridging the gap between simulated testing and safe real-world deployment.

\clearpage
\appendix
\section{Contributors and Acknowledgments}
\label{sec::contributors}

\subsection{Contributors}
Research contributors (alphabetically): Aarti Basant, Amlan Kar, Despoina Paschalidou, Fangyin Wei, Francesco Ferroni, Guillermo Garcia Cobo, Haithem Turki, Huan Ling, Jaewoo Seo, James Lucas, Jay Zhangjie Wu, Jialiang Wang, Jonathan Lorraine, Jun Gao, Kai He, Katarina Tothova, Kevin Xie, Michał Tyszkiewicz, Qi Wu, Riccardo de Lutio, Ruilong Li, Sanja Fidler, Seung Wook Kim, Tianchang Shen, Tianshi Cao, Tobias Pfaff, William Lew, Xindi Wu, Xuanchi Ren, Yifan Lu, Yuxuan Zhang, Zan Gojcic, Zian Wang.

\subsection{Acknowledgements}
We would also like to additionally acknowledge (alphabetically): Aditya Mahajan, Andras Bodis-Szomoru, Andy Ju, Arnav Khanna, Ashley Goldstein, Bartosz Stefanik, Boris Ivanovic, Bruno Costa Rendon, Cliff Woolley, Gangzheng Tong, Jeff Pei, Jesse Archer, Jonathan McCaffrey, Maciej Bala, Marco Pavone, Martin Ding Ma, Matt Cragun, Maximilian Igl, Michael Watson, Ming-Yu Liu, Nat Duca, Peter Karkus, Pyarelal Knowles, Ross Luo, Wonsik Han, Yan Wang, Yong He for their invaluable contributions in the integration of Alpamayo 1 in a closed-loop with OmniDreams, as well as open sourcing OmniDreams.

\clearpage
\setcitestyle{numbers}
\bibliographystyle{plainnat}
\bibliography{main}

\clearpage
\section{Glossary}
\label{sec::glossary}

This whitepaper draws on terminology from autonomous-driving research, video diffusion modeling, parallel-attention systems, and the NVIDIA SIL platform stack. Acronyms and product names are listed below for reference; each term is also expanded on first mention in the body.

\subsection*{Concepts and acronyms}

\begin{description}
\setlength{\itemsep}{0pt}
\item[\textbf{AV}] autonomous vehicle.
\item[\textbf{VLA}] vision-language-action model. A multimodal foundation model that ingests images or video plus text and emits an action (a planned trajectory or control signal).
\item[\textbf{WAM}] World-Action Model. A video-conditioned policy that maps video input to actions without an explicit language modality.
\item[\textbf{VLM}] vision-language model.
\item[\textbf{BEV}] bird's-eye view; the top-down 2D ego-centric scene representation used by perception and trajectory pipelines.
\item[\textbf{HD map}] high-definition map; vector representation of lanes, signs, crosswalks, and dynamic-agent bounding boxes used here as a structured conditioning signal.
\item[\textbf{VRU}] vulnerable road user; pedestrians, cyclists, scooter riders, and similar non-vehicle agents.
\item[\textbf{LiDAR}] light detection and ranging; an active depth-sensing modality, referenced here in the context of neural-reconstruction baselines.
\item[\textbf{DiT}] diffusion transformer; the architecture family that backs \modelnamesv and \modelnamemv.
\item[\textbf{VAE}] variational autoencoder; the tokenizer that maps between pixels and the DiT's latent space.
\item[\textbf{MLP}] multi-layer perceptron; used for the lightweight control-token encoder and the per-layer projections in the DiT.
\item[\textbf{AdaLN}] adaptive layer normalization; conditioning is injected as an additive shift, scale, and gate modulation of normalized activations.
\item[\textbf{RoPE}] rotary position embedding; relative positional encoding applied in the attention layers, with frequencies shared across denoising steps.
\item[\textbf{KV cache}] key-value cache used by autoregressive transformers to amortize attention across steps; central to streaming inference (\cref{sec::inf_kvcache}).
\item[\textbf{Streaming KV cache}] a bounded, in-place KV cache with static shapes that supports CUDA-graph capture and rolling-window eviction.
\item[\textbf{Attention-sink tokens}] a small fixed prefix of tokens kept in the KV cache to preserve long-range stability under local-window attention.
\item[\textbf{Local-window attention}] temporal attention restricted to a fixed-length recent window of chunks, giving constant per-chunk attention cost.
\item[\textbf{Cross-view attention}] attention applied across simultaneously-generated camera views at each time step to enforce inter-view consistency.
\item[\textbf{Context parallelism}] sharding strategy that splits a sequence across GPUs along the spatial, temporal, or view axis; \cref{sec::inf_mgpu} introduces a hierarchical $V\!\times\!T\!\times\!HW$ factorization.
\item[\textbf{Ring attention}] a context-parallel attention kernel that pipelines key-value shard exchanges against the local attention computation~\citep{liu2024ringattention}.
\item[\textbf{Diffusion Forcing}] a training paradigm in which each token receives an independent random noise level, enabling a single model to act both as a next-token predictor and as a full-sequence denoiser~\citep{chen2024diffusionforcing}.
\item[\textbf{Self Forcing}] a training paradigm that closes the autoregressive train-test gap by performing self-rollout during training, conditioning each frame on its own previously-generated outputs~\citep{huang2025self}.
\item[\textbf{DMD}] Distribution Matching Distillation~\citep{yin2024one}; the holistic, video-level distribution-matching objective used in our distillation stage.
\item[\textbf{KL}] Kullback--Leibler divergence; the distillation objective minimizes a (reverse) KL divergence between the model and data distributions.
\item[\textbf{Rectified flow}] training objective used by the base Cosmos backbone~\citep{liu2022flow,nvidia2025worldsimulationvideofoundation}.
\item[\textbf{Pixel-shuffle (HD map encoder)}] a parameter-free patch-wise encoder that produces tokens at the same spatial and temporal resolution as the RGB branch, replacing a learned 3D VAE for the HD map signal.
\item[\textbf{TAE}] a fast VAE decoder from LightTAE.
\item[\textbf{CUDA Graphs}] an NVIDIA execution-model construct that lets the entire DiT forward pass be replayed as a single GPU submission~\citep{nvidia_cuda_graphs}.
\item[\textbf{Pre-fetch / post-fetch}] two strategies for integrating chunk-based generation with closed-loop simulation; \cref{sec::applications} adopts pre-fetch for AlpaSim.
\item[\textbf{gRPC, NCCL, RDMA}] communication primitives used by AlpaSim for cross-process coordination, by the multi-GPU video model for collective ops, and as a future direction for high-throughput frame transport.
\item[\textbf{ControlNet}] a control-adapter pattern for diffusion models, included here as a contrast to the lightweight control branch we use.
\item[\textbf{FVD}] Fr\'echet Video Distance; a distribution-level video-quality metric referenced in \cref{sec::results}.
\item[\textbf{LET-AP / LET-APL / LET-APH}] Longitudinal-Error-Tolerant Average Precision and its localization- (APL) and heading-weighted (APH) variants; 3D object-detection metrics reported in \cref{sec::results}, computed with an off-the-shelf detector~\citep{hung2024}.
\end{description}

\subsection*{NVIDIA SIL platform components}

\begin{description}
\setlength{\itemsep}{0pt}
\item[\textbf{SIL}] NVIDIA Spatial Intelligence Lab.
\item[\textbf{Cosmos / Cosmos-Predict 2.5}] NVIDIA's video-foundation-model platform~\citep{nvidia2025worldsimulationvideofoundation}; \modelnamesv and \modelnamemv post-train from Cosmos-Predict\,2.5.
\item[\textbf{OmniDreams}] the system documented in this whitepaper.
\item[\textbf{\modelnamesv\ / \modelnamemv}] the single-view and multi-view variants described in \cref{sec::model}.
\item[\textbf{Cosmos-Drive-Dreams}] NVIDIA's synthetic-driving-data generation pipeline~\citep{ren2025cosmosdrivedreams}, complementary to OmniDreams.
\item[\textbf{Alpamayo / Alpamayo 1}] NVIDIA's open-source AV policy stack~\citep{nvidia2026alpamayo}; the policy that drives the closed-loop demos in \cref{sec::applications}.
\item[\textbf{AlpaSim}] NVIDIA's open-source closed-loop AV simulator~\citep{nvlabs2025alpasim}; the orchestrator into which OmniDreams plugs in \cref{sec::applications}.
\item[\textbf{NuRec / Omniverse NuRec}] NVIDIA's neural-reconstruction tool for AV simulation~\citep{nvidia2024nurec}; the reconstruction-based renderer that NuRec replaces in our integration.
\item[\textbf{SIL-Wheel}] NVIDIA SIL's multi-modal AV video search and curation platform~\citep{nvidia2026silwheel}; used to construct training and evaluation slices for the experiments in \cref{sec::data,sec::results}. See \cref{sec::silwheel}.
\item[\textbf{GB300}] NVIDIA Grace Blackwell platforms; \cref{sec::inference} reports per-chunk latency on a single GB300 and a $16$-GPU GB300 cabinet.
\end{description}

\subsection*{Datasets and benchmarks}

\begin{description}
\setlength{\itemsep}{0pt}
\item[\textbf{RDS}] the multi-view AV training corpus used for mid-training; sourced from real-world driving logs at $1080$p, $30$\,FPS across $15$ countries (\cref{sec::data_sources}).
\item[\textbf{RDS-HQ-1M}] the post-training corpus, with quality- and HD map-focused human verification.
\end{description}

\subsection*{Mathematical notation}

The following symbols recur throughout the paper. A few letters ($T$, $K$, $L$) are reused with different meanings across sections; the per-section reuse is noted explicitly below.

\paragraph{Latent variables.}
\begin{description}
\setlength{\itemsep}{0pt}
\item[$\mathrm{x}$] a latent video, the output of the Cosmos VAE encoder; the variable the diffusion process operates on.
\item[$\mathrm{x}^{1:T}$] a latent sequence of $T$ frames. $\mathrm{x}^i$ is the $i$-th frame, $\mathrm{x}^{<i}$ the autoregressive prefix.
\item[$\mathrm{x}_t$] the noisy latent at flow-matching time $t \in [0,1]$, defined by the linear interpolation $\mathrm{x}_t = (1-t)\,\mathrm{x} + t\,\epsilon$.
\item[$\mathrm{t}$] per-token timestep vector under Diffusion Forcing; each token receives its own $t$ sampled independently.
\item[$\hat{x}$] a sample produced by the model (denoised latent or self-rollout output, depending on context).
\end{description}

\paragraph{Noise, conditioning, and parameters.}
\begin{description}
\setlength{\itemsep}{0pt}
\item[$\epsilon$] Gaussian noise, $\epsilon \sim \mathcal{N}(0, I)$.
\item[$\mathrm{c}$] auxiliary conditioning: text caption, first-frame image, HD map control, ego trajectory.
\item[$\theta$] learnable parameters of the OmniDreams DiT (and, where indicated, of distilled variants).
\end{description}

\paragraph{Distributions and operators.}
\begin{description}
\setlength{\itemsep}{0pt}
\item[$\mathcal{N}(0, I)$] standard multivariate normal.
\item[$\mathrm{Uniform}(1,\ldots,T)$] discrete uniform distribution on the timestep index used for Self Forcing backprop.
\item[{$\mathbb{E}_{\,\cdot\,}[\cdot]$}] expectation, with the variables averaged over listed in the subscript.
\item[{$\mathcal{O}(\cdot)$}] asymptotic complexity in the indicated arguments.
\item[{$\mathrm{sg}[\cdot]$}] stop-gradient operator (treat the argument as a constant for backprop).
\item[{$\|\cdot\|^2$}] squared Euclidean norm.
\end{description}

\paragraph{Model functions and distributions.}
\begin{description}
\setlength{\itemsep}{0pt}
\item[$\mathbf{u}_\theta(\mathrm{x}_t, t)$] rectified-flow velocity-prediction network. $\mathbf{u}_\theta(\mathrm{x}_t, t;\mathrm{c})$ is the conditional variant.
\item[$u_\theta(\mathrm{x}^i_\mathrm{t} \mid \mathrm{x}^{<i})$] causal/autoregressive velocity predictor over a latent prefix (\cref{sec::training}).
\item[$\mathrm{v}_t$] flow-matching velocity target, $\mathrm{v}_t = \epsilon - \mathrm{x}$.
\item[$p_\theta,\ p_\mathrm{data}$] model and data distributions; the distillation objective minimizes a divergence between them.
\item[$\mathbf{f}_\phi,\ \mathbf{f}_\psi$] score networks in Distribution Matching Distillation: $f_\phi$ is a frozen real-data score; $f_\psi$ is a learned fake-data score. The update direction is $f_\psi - f_\phi$ (\cref{sec::training}).
\item[$\mathbf{L}$] (training-section meaning) the training loss, e.g. $\mathbf{L}_{DF} = \mathbb{E}_{\mathrm{x}^{1:T}, \epsilon} \left[ \|\mathbf{u}\theta(\mathrm{x}^{1:T}_\mathrm{t}, \mathrm{t} ) -  \mathrm{v}_\mathrm{t}\|^2\right]$ for Diffusion Forcing training.
\end{description}

\paragraph{Sizes and parallelism axes.}
\begin{description}
\setlength{\itemsep}{0pt}
\item[$T$] (sequence meaning) number of latent frames in a generation chunk or sequence. \emph{Overload:} in \cref{sec::inf_mgpu}, $T$ denotes the temporal-axis context-parallelism degree.
\item[$K$] (inference meaning) frames per generation chunk; production multi-view configuration uses $K{=}16$. \emph{Overload:} in \cref{sec::training}, $K$ is the number of denoising steps in the Self Forcing self-rollout, with $K{=}2$ at the timestep schedule $[1000, 450]$.
\item[$N$] number of camera views in multi-view generation; production deployment uses $N{=}4$ (front-wide, left, right, front-telescope).
\item[$V,\ T,\ HW$] degrees of context parallelism along the view, temporal, and spatial axes, factorized as $V\!\times\!T\!\times\!HW$ in \cref{sec::inf_mgpu}.
\item[$L$] (inference meaning) rolling-window size: in \cref{sec::training} it bounds the rolling KV cache (cache of the most recent $L$ frames), and in \cref{sec::inf_local_attn} it bounds the local temporal-attention window measured in chunks.
\item[$S,\ s$] $S$ is the total number of denoising steps used by a checkpoint; $s$ is a step index sampled per training iteration for Self Forcing backprop, $s \sim \mathrm{Uniform}(1,\ldots,T)$.
\end{description}

\paragraph{Symbols in tables and prose.}
\begin{description}
\setlength{\itemsep}{0pt}
\item[$\uparrow\,/\,\downarrow$] in result tables, higher-is-better / lower-is-better for the column metric.
\item[$\rightarrow$] before/after, typically reporting latency or quality under a single stack-level optimization (e.g.\ $1520\,\text{ms} \rightarrow 1166\,\text{ms}$).
\item[$\times$] cartesian product (e.g.\ in resolution $704{\times}1280$) and multiplicative scaling (e.g.\ $1.5\times$ speedup).
\end{description}

\end{document}